\documentclass[letterpaper]{article} 
\usepackage{aaai2026}  
\nocopyright

\usepackage{times}  
\usepackage{helvet}  
\usepackage{courier}  
\usepackage[hyphens]{url}  
\usepackage{graphicx} 
\usepackage{natbib}  
\usepackage{caption} 
\usepackage{url} 
\usepackage{booktabs} 
\usepackage{amsfonts,amsmath,amssymb,amsthm} 
\usepackage{microtype} 
\usepackage{graphicx} 
\usepackage{tikz} 
\usepackage{xspace} 

\theoremstyle{definition}
\newtheorem{definition}{Definition}

\newcommand{\refsec}[1]{Section~\ref{#1}}
\newcommand{\reffig}[1]{Figure~\ref{#1}}
\newcommand{\reftab}[1]{Table~\ref{#1}}
\newcommand{\refappsec}[1]{Appendix Section~\ref{#1}}
\newcommand{\refappfig}[1]{Appendix Figure~\ref{#1}}

\newcommand{\refappdef}[1]{Appendix Definition~\ref{#1}}

\newcommand{\DTNet}{{\tt DT-Net}\xspace}
\newcommand{\PINet}{{\tt PI-Net}\xspace}
\newcommand{\ITNet}{{\tt IT-Net}\xspace}
\newcommand{\FFNet}{{\tt FF-Net}\xspace}

\title{On Logical Extrapolation for Mazes 
\\ with Recurrent and Implicit Networks}
\author{
    Brandon Knutson\textsuperscript{\rm 1},
    Amandin Chyba Rabeendran\textsuperscript{\rm 2},
    Michael Ivanitskiy\textsuperscript{\rm 1}, \\
    Jordan Pettyjohn\textsuperscript{\rm 1},
    Cecilia Diniz Behn\textsuperscript{\rm 1},
    Samy Wu Fung\textsuperscript{\rm 1}, 
    Daniel McKenzie\textsuperscript{\rm 1},
}
\affiliations{
    \textsuperscript{\rm 1}Department of Applied Mathematics and Statistics, Colorado School of Mines\\
    \texttt{\{bknutson,mivanits,jpettyjohn,cdinizbe,dmckenzie,swufung\}@mines.edu}\\
    \textsuperscript{\rm 2}Courant Institute of Mathematical Sciences, New York University\\
    \texttt{atc9782@nyu.edu}
}

\begin{document}

\maketitle

\begin{abstract}
Recent work suggests that certain neural network architectures---particularly recurrent neural networks (RNNs) and implicit neural networks (INNs)---are capable of logical extrapolation. When trained on easy instances of a task, these networks (henceforth: logical extrapolators) can generalize to more difficult instances. Previous research has hypothesized that logical extrapolators do so by learning a scalable, iterative algorithm for the given task which converges to the solution. We examine this idea more closely in the context of a single task: maze solving. By varying test data along multiple axes --- not just maze size --- we show that models introduced in prior work fail in a variety of ways, some expected and others less so. It remains uncertain whether any of these models has truly learned an algorithm. However, we provide evidence that a certain RNN has approximately learned a form of `deadend-filling'. We show that training these models on more diverse data addresses some failure modes but, paradoxically, does not improve logical extrapolation. We also analyze convergence behavior, and show that models explicitly trained to converge to a fixed point are likely to do so when extrapolating, while models that are not may exhibit more exotic limiting behavior such as limit cycles, {\em even when} they correctly solve the problem. Our results (i) show that logical extrapolation is not immune to the problem of {\em goal misgeneralization}, and (ii) suggest that analyzing the {\em dynamics} of extrapolation may yield insights into designing better logical extrapolators.
\end{abstract}


\section{Introduction}
\label{sec:introduction}
A key feature of human learning is the ability to generalize from easy problem instances to harder ones, often by thinking for longer \citep{schwarzschild2021can}. Work in multiple areas of machine learning has provided empirical evidence that neural networks are also capable of such {\em logical extrapolation} (also known as upwards generalization or algorithmic reasoning \citep{velivckovic2022clrs}). Logical extrapolation is a form of out-of-distribution (OOD) generalization in which the test distribution is shifted away from the training distribution along an intuitively defined difficulty axis. These networks usually utilize {\em test-time scaling}, which increases their computational budget at test time to boost inference \citep{dehghani2018universal,gilton2021deep,schwarzschild2021can,bansal2022endtoend,anil2022path,openai2024reasoning,geiping2025scaling}. In this work we conduct an in-depth study of logical extrapolation for two classes of neural networks that exhibit test-time scaling by iterating a recurrent block additional times: weight-tied RNNs and INNs, also known as Deep Equilibrium Networks (DEQs) \citep{bai2019deep,el2021implicit,fung2022jfb}. We focus on a single task --- maze-solving --- a canonical spatial-reasoning task. We are interested in two intertwined questions:
\begin{enumerate}
    \item[Q1] \textit{Do RNNs and INNs learn an algorithm for maze-solving?}
    \item[Q2] \textit{Are the recurrent iterations converging to something? If yes, to what? Does this affect logical extrapolation in any way?}
\end{enumerate}

\begin{figure}
    \centering
    \includegraphics[width=1.0\linewidth]{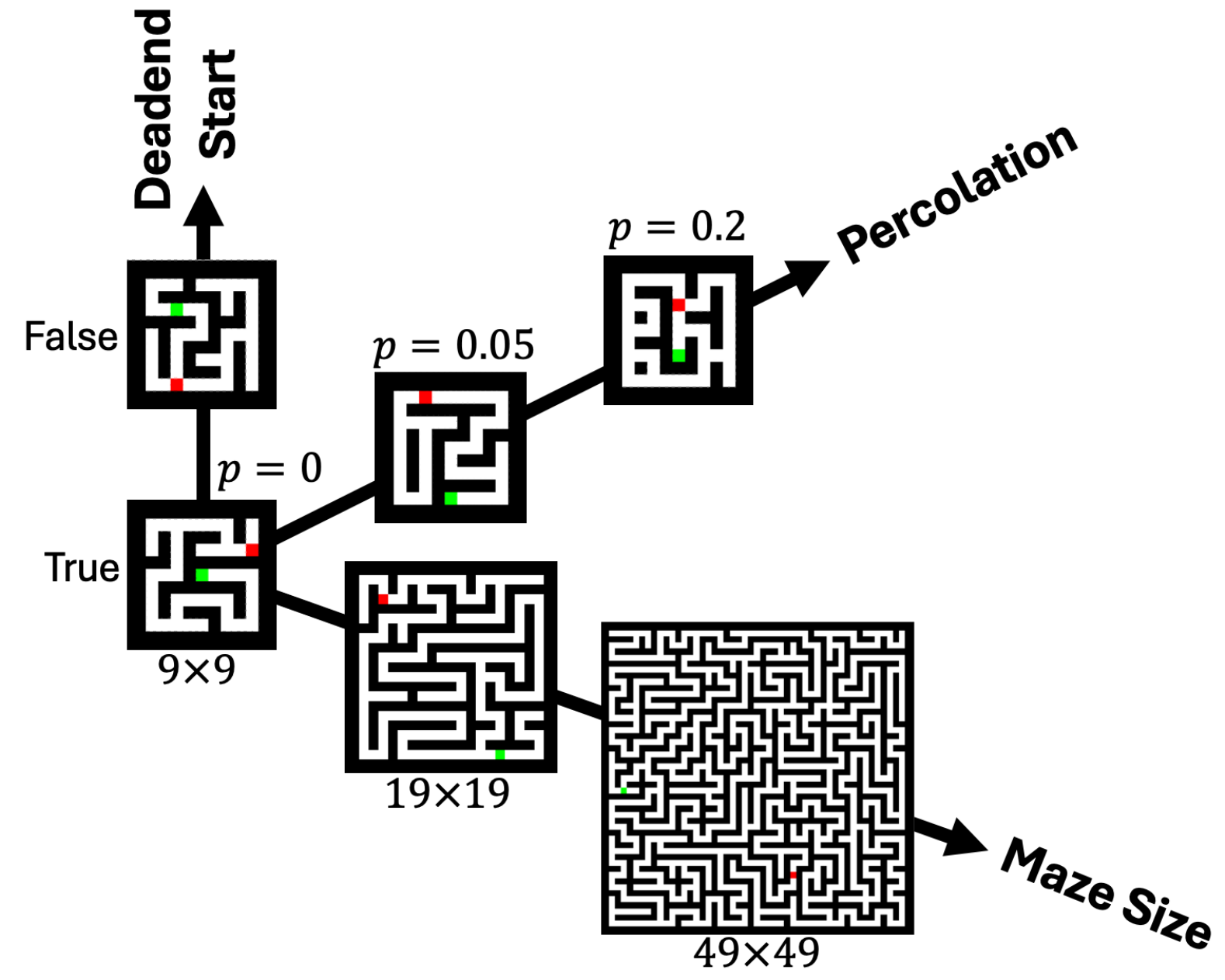}
    \caption{
        Three extrapolation dimensions: maze size, percolation, and deadend start. Each shown maze is generated using the indicated parameters.  Green denotes the start position and red denotes the target.
    }
    \label{fig:maze_extrap_visual}
\end{figure}

We are motivated by various intriguing observations and partial results in the recent literature \cite{anil2022path,bansal2022endtoend,schwarzschildalgorithm,schwarzschild2021can,schwarzschild2021thinking}. Answering Q1 is necessary for assessing the robustness and trustworthiness of logical extrapolators, as we take "learn an algorithm" to mean "learn a correct and interpretable algorithm that scales to harder problems". Q2 is important philosophically, as many prior works explain the success of logical extrapolators by claiming that the recurrent part converges to a fixed point. While we confine our analysis to a single task, we suspect our insights will transfer to similar tasks.

To tackle Q1, we use the {\tt maze-dataset} package \cite{ivanitskiy_maze-dataset_2023,ivanitskiy2023maze}. While previous studies on logical extrapolation in maze-solving have varied the test distribution by increasing maze size only, {\tt maze-dataset} allows two new distribution shifts: (i) a variable $\texttt{deadend\_start} \in \{\texttt{True},\texttt{False}\}$ which, when set to \texttt{True}, constrains the start point to have exactly one neighbor, and (ii) a percolation constant $p \in [0,1]$ which probabilistically determines the number of cycles in the maze. See Figure~\ref{fig:maze_extrap_visual}. Prior work \citep{bansal2022endtoend, anil2022path} has set $\texttt{percolation}=0$ and $\texttt{deadend\_start}=\texttt{True}$ for both training and testing. By varying $p$ and \texttt{deadend\_start} we expose pretrained models (the RNN of \cite{bansal2022endtoend} and the INN of \cite{anil2022path}) to corner cases not considered before. Examining how both models handle mazes containing cycles provides additional evidence that the RNN has learned a variant of the deadend-filling algorithm \citep{hendrawan2020deadend-fill}, bolstering results of \cite{schwarzschildalgorithm}. However, by toggling the seemingly inconsequential $\texttt{deadend\_start}$ variable, we produce examples of mazes where deadend-filling would succeed, yet both models fail, suggesting that the RNN has {\em not} learned deadend-filling. Thus, our answer to Q1 is:

\begin{enumerate} 
    \item[A1] \textit{Partially. One RNN approximately learns deadend-filling — it succeeds like deadend-filling on acyclic mazes and fails like deadend-filling on mazes with cycles\footnote{Specifically, the output of the RNN agrees with deadend-filling on 98.8\% of 143k tests mazes across a variety of sizes and percolation values.}. However, it also fails on mazes without a deadend start, revealing a mismatch. We trained many other models, but were unable to find clear-cut evidence of algorithm learning.}
\end{enumerate}

 While \citep{bansal2022endtoend} observes that, with proper training, RNNs converge to a fixed point even when extrapolating, the results of \citep{anil2022path} hint at more complex behavior. Specifically, \citet[App. F, App. G]{anil2022path} find evidence that RNNs may sometimes converge to limit cycles. To address Q2 rigorously, we quantify this phenomenon using tools from {\em Topological Data Analysis} (TDA) 
 \citep{skraba2012topological,perea2015sliding,tralie2018quasi}.  We find that, while the INN from \citep{anil2022path} consistently converges to a fixed point, regardless of maze size, the RNN from \citep{bansal2022endtoend} exhibits more complex limiting behavior. To be correct, the iterative part of a logical extrapolator must converge into the pre-image of the solution after sufficient iterations. However, the exact convergence pattern inside the pre-image does not appear to matter --- logical extrapolators that sometimes converge to limit cycles are as performant as extrapolators that always converge to fixed points. Thus, our answer to Q2 is:
 \begin{enumerate}
     \item[A2] \textit{Yes. Sometimes a fixed point, sometimes something more exotic. No, it doesn't seem to matter.}
 \end{enumerate}

It is perhaps unsurprising that the pre-trained models we consider \cite{bansal2022endtoend,anil2022path} do not generalize to mazes with cycles, as they did not see such mazes in their training data. By retraining them using more diverse training data (i.e., $p > 0$) we investigate whether these models can be made more performant. We observe that a little bit goes a long way: if even a tiny fraction of training mazes contain cycles, performance is dramatically improved. We hypothesize that this may be because the models have learned to emulate a different maze solving algorithm---one which {\em does generalize} to mazes with cycles---although identifying this algorithm remains challenging. 

\section{Preliminaries and related works}
\label{sec:preliminaries}

\subsection{The maze-solving task}
\label{sec:maze-solving_task}

\begin{figure}[t]
    \centering
    \includegraphics[width=0.44\linewidth]{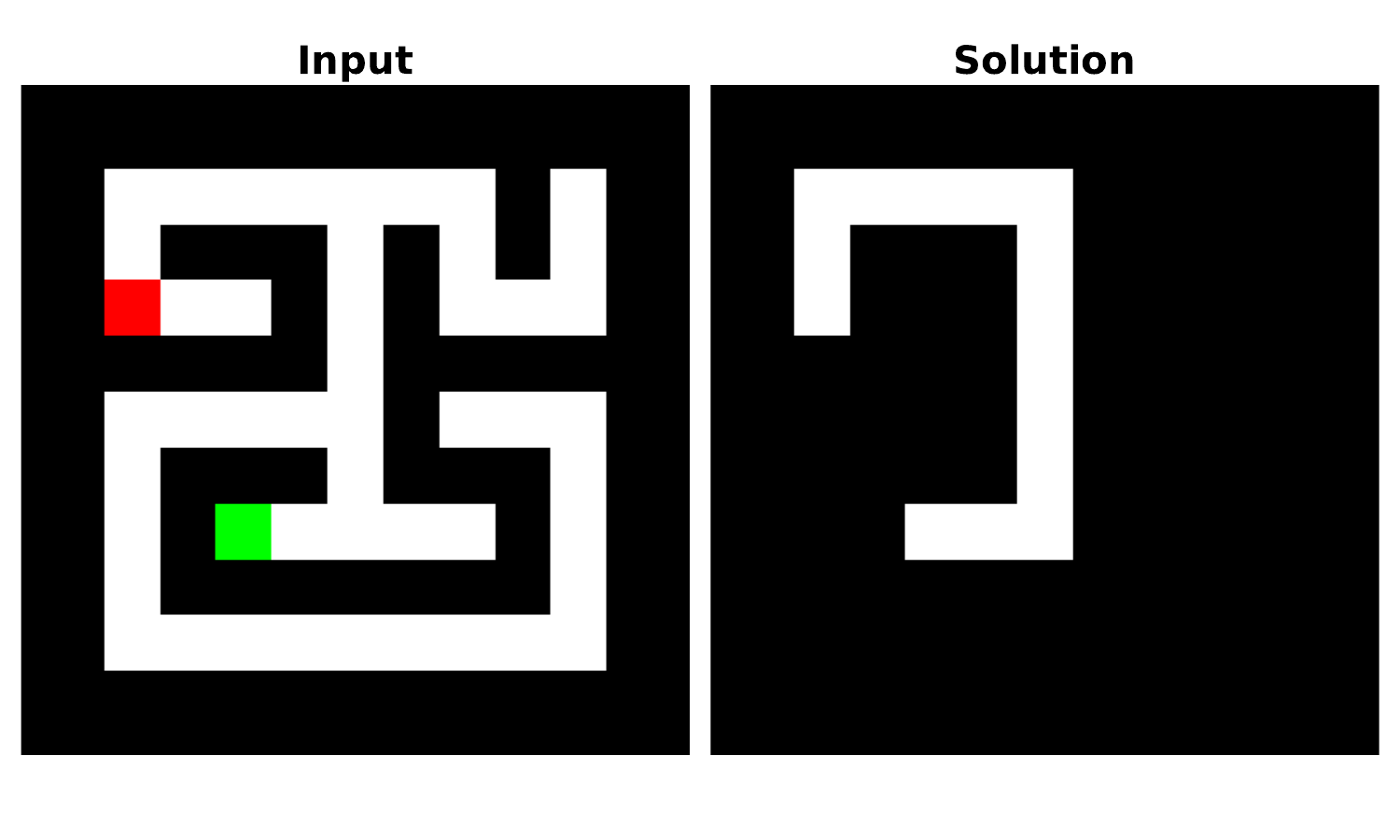}
    \includegraphics[width=0.44\linewidth]{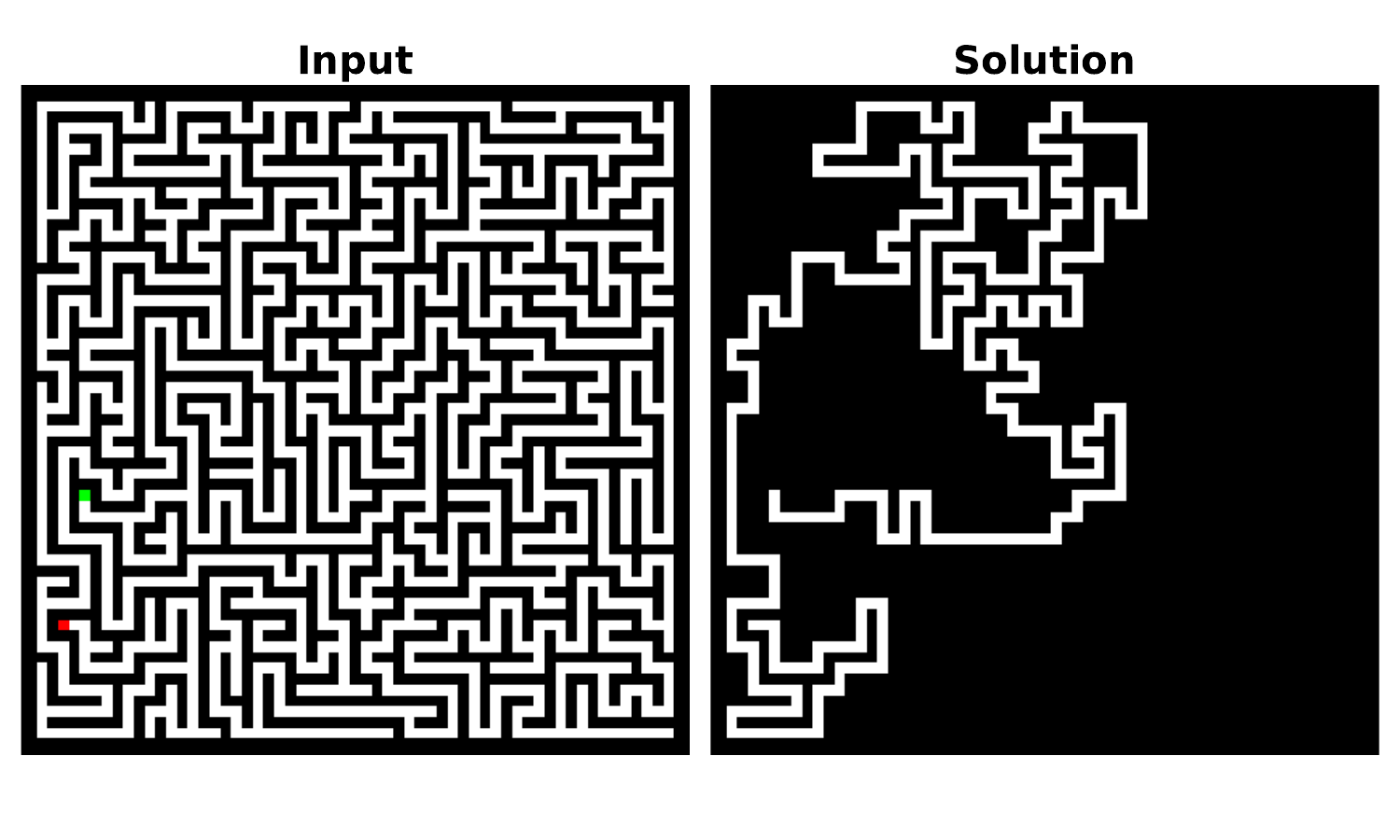}
    \caption{
        Maze input-solution pairs of size $9 \times 9$ (left) and $49 \times 49$ (right). Start positions are in \textbf{green} and end positions are in \textbf{red}. Mazes problems/inputs are RGB raster images and solutions are black and white images highlighting the solution path in white.
    }
    \label{fig:maze_examples}
\end{figure}

In this and prior work \citep{schwarzschild2021datasets,schwarzschild2021can,bansal2022endtoend,anil2022path}, maze solving problems are encoded as raster images (\reffig{fig:maze_examples}). Given this RGB input image, the task is to return a black and white image representing the unique path from start to end. We consider the ``accuracy'' of a model on a single maze to be $1$ if its prediction is a valid path of minimal length and 0 otherwise. Instead of using the original ``easy to hard'' dataset \citep{schwarzschild2021datasets}, we use the {\tt maze-dataset} package \citep{ivanitskiy_maze-dataset_2023} which provides the flexibility to modify the distribution from which (maze, solution) pairs are drawn. The original ``easy to hard'' dataset only contains acyclic (i.e. $\texttt{percolation}=0$, any path between two nodes is unique) mazes generated via randomized depth-first search (RDFS) and with start positions having exactly degree 1 (i.e., {\tt deadend\_start=True}). We remove these restrictions to investigate the behavior of the selected models on out-of-distribution mazes in \refsec{sec:maze-solving_task}. More details on our usage of {\tt maze-dataset} are given in \refappsec{app:maze-dataset}.

\begin{figure}[t]
    \centering
    \includegraphics[width=0.44\linewidth]{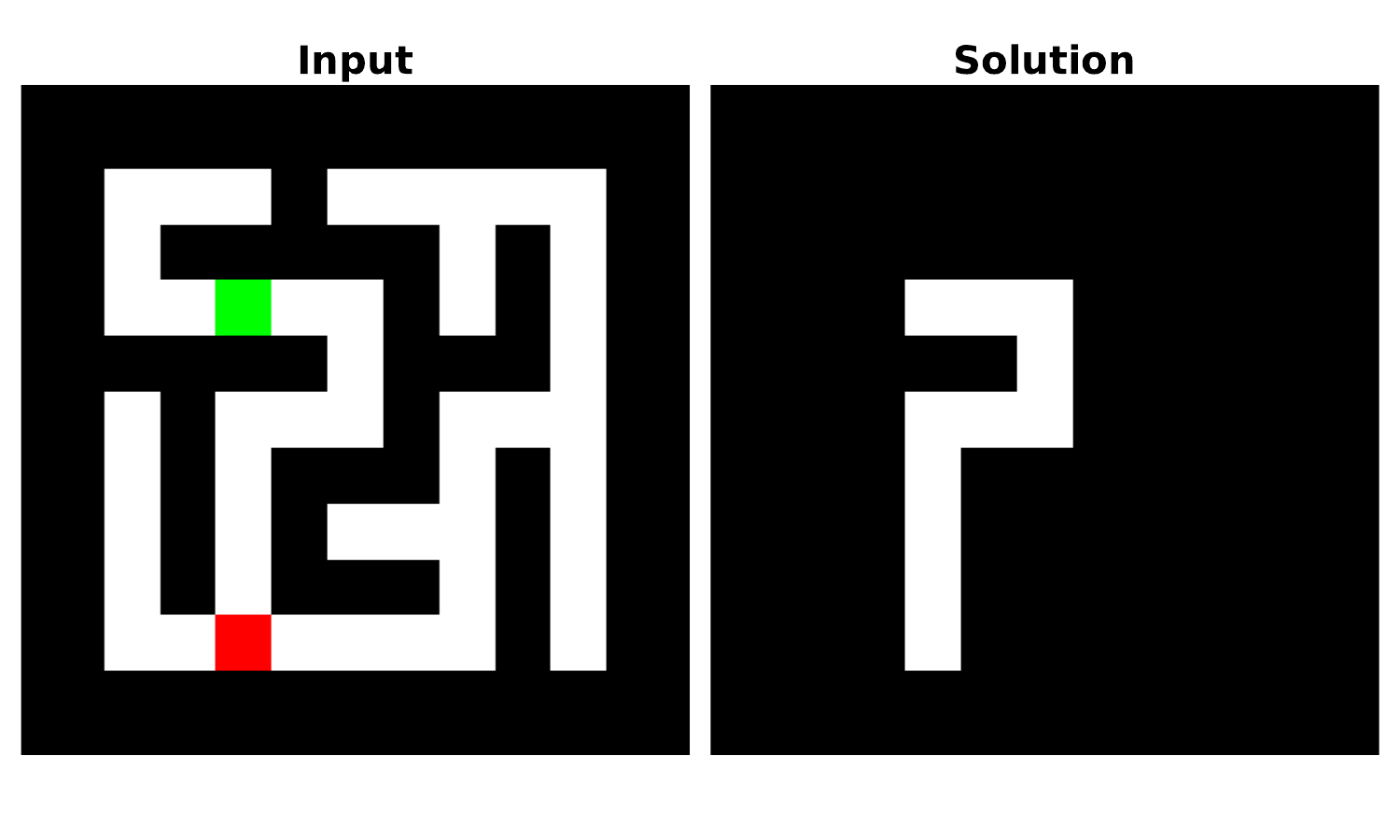}
    \includegraphics[width=0.44\linewidth]{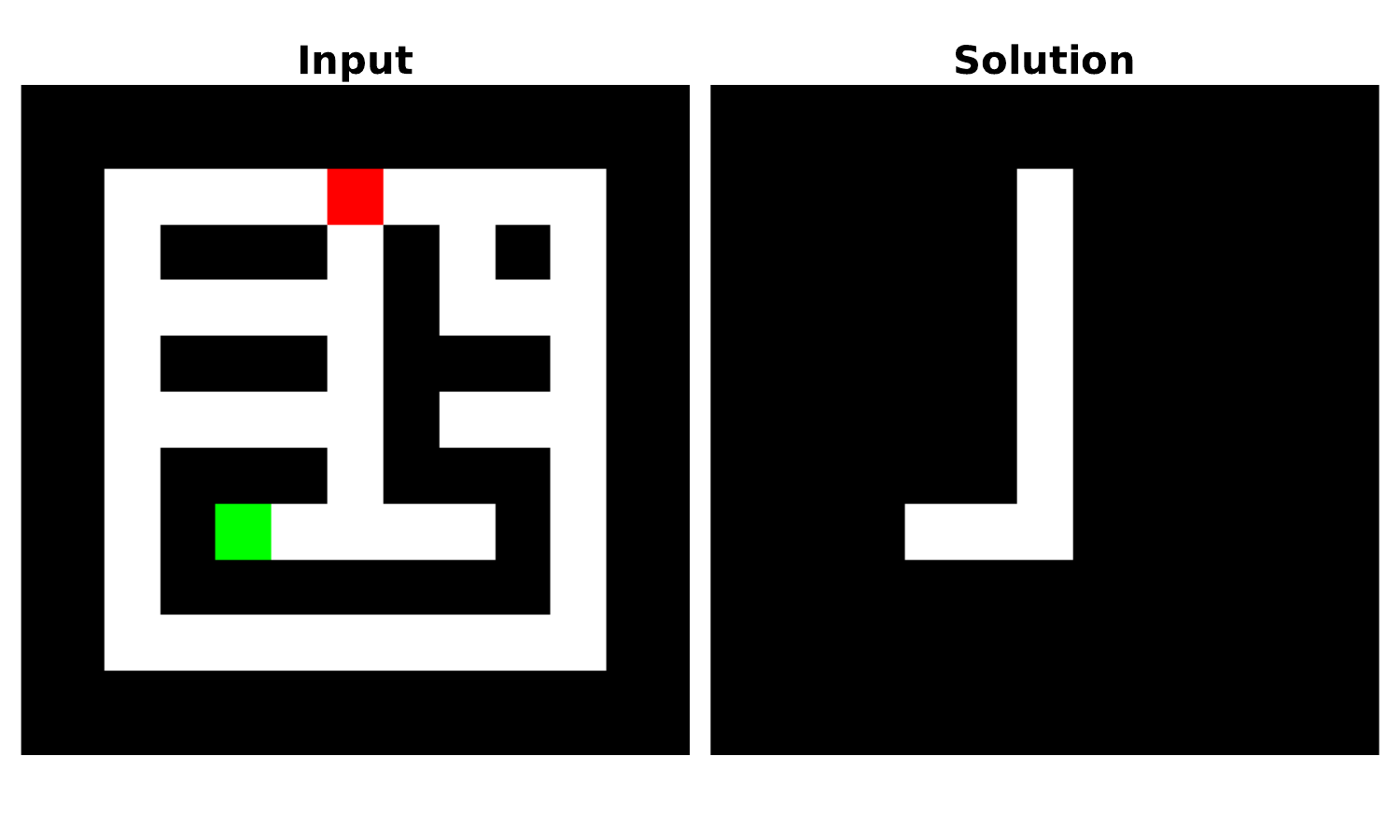}
    \caption{
        Example of maze with a start position that has multiple neighbors (left); example of a percolated maze ($\texttt{percolation}=0.2$) with loops (right).
    }
    \label{fig:maze_examples_deadend_percolation}
\end{figure}
\subsection{Recurrent neural networks}
\label{sec:RNNs}

A special class of RNNs, namely, weight-tied input-injected networks (or simply weight-tied RNNs), are used in logical extrapolation~\citep{schwarzschild2021can,bansal2022endtoend}. For a $K$-layer weight-tied RNN $\mathcal{N}_\Theta$, the output is given by
\begin{equation}
    \begin{split}
    \mathcal{N}_\Theta(d) &= P_{\Theta_2}(u_K), \\
    \text{where } u_{j} &= T_{\Theta_1}(u_{j-1}, d)\quad\text{for }j=1,\ldots,K.
    \end{split}
\end{equation}
Here, $\Theta_1$ and $\Theta_2$ are the parameters of the networks $T_{\Theta_1}$ and $P_{\Theta_2}$ respectively, and $\Theta := \{\Theta_1,\Theta_2\}$, while $d$ denotes the input features. These networks represent a unique class of architectures that leverage weight sharing across layers to reduce the number of parameters. 
The input injection at each layer ensures the network does not `forget' the initial data \citep{bansal2022endtoend}. In~\citep{bansal2022endtoend}, it is empirically observed that a certain weight-tied RNN extrapolates to larger mazes when applying more iterations. The authors speculate that the reason for this success is that the model has learned to converge to fixed points within its latent space (\refsec{sec:conclusion}, \cite{bansal2022endtoend}).

\subsection{Implicit networks}
\label{sec:INNs}

Drawing motivation from~\citep{bansal2022endtoend}, \citet{anil2022path} propose to use implicit neural networks (INNs) for logical extrapolation tasks.
INNs are a broad class of architectures whose outputs are the fixed points of an operator parameterized by a neural network. That is,
\begin{equation}
    \mathcal{N}_\Theta(d) = P_{\Theta_2}(u_\star) \quad \text{ where } \quad u_\star = T_{\Theta_1}(u_\star, d).
\end{equation}
Here again $\Theta = \{\Theta_1,\Theta_2\}$ refers collectively to the parameters of the networks $T_{\Theta_1}$ and $P_{\Theta_2}$ and $d$ is the input feature, while $u_\star$ represents a fixed point of $T_\Theta$. These networks can be interpreted as infinite-depth weight-tied input-injected recurrent neural networks~\citep{el2021implicit, bai2019deep, winston2020monotone, fung2024generalization}. Unlike traditional networks, INN outputs are not defined by a fixed number of computations but rather by an implicit condition. INNs have been applied to domains as diverse as image classification \citep{bai2020multiscale}, inverse problems \citep{gilton2021deep, Yin2022Learning, liu2022online, heaton2021feasibility, heaton2023explainable}, optical flow estimation \citep{bai2022deep}, game theory \citep{mckenzie2024three}, and decision-focused learning~\citep{mckenzie2024differentiating}. In principle, INNs are naturally suited for logical extrapolation tasks for which solutions can be characterized by a fixed point condition. Key to logical extrapolation is that this characterization is always the same, regardless of the difficulty of the problem at hand.

\subsection{Topological data analysis in the latent space}
\label{sec:TDA}

For both RNNs and INNs, we call $\{u_j\}_{j=1}^{K}\subset \mathbb{R}^n$ the {\em latent iterates}, and $n$ the latent dimension. Note that $n$ may be significantly larger than the output space dimension, so $P_{\Theta_2}$ is a projection operator with large-dimensional fibers. 
Although models are trained to reduce loss (i.e. the discrepancy between $\mathcal{N}_{\Theta}(d)$ and the true solution $x^{\star}$) at every iteration \citep{bansal2022endtoend,anil2022path,schwarzschild2021can}, there is no incentive for the iterative part of the network $T_{\Theta_1}(\cdot, d)$ to prefer one element of the fiber $P^{-1}_{\Theta_2}(x^{\star}) := \{u \in \mathbb{R}^n: P_{\Theta_2}(u) = x^{\star}\}$ over another, unless additional constraints are imposed upon $T_{\Theta_1}$, such as contractivity \cite{el2021implicit} or monotonicity \citep{winston2020monotone}. Thus,  $T_{\Theta_1}(\cdot, d)$ may exhibit more complex dynamics than convergence-to-a-point, while $\mathcal{N}_{\Theta}(d)$ still yields the correct solution. \citet{anil2022path} proposed that, in order for $T_{\Theta_1}(\cdot, d)$ to exhibit logical extrapolation\footnote{Note they call this property `upwards generalization'.}, it need not have a unique fixed point, but rather need only possess a global attractor. In other words, no matter which initialization $u_{0}$ is selected, the latent iterates should exhibit the same asymptotic behaviour. They dub this property ``path independence''. In \citep[App. F, App. G]{anil2022path} evidence is provided of instances $d$ where the latent iterates induced by $T_{\Theta_1}(\cdot, d)$ form a limit cycle, yet $\mathcal{N}_{\Theta}(d)$ is correct. 

If RNNs and INNs can generalize while converging to limit cycles, this has consequences for the design of logical extrapolators. For example, it is common practice to terminate the iteration when the residuals $r_j := \|u_{j+1} - u_j\|_2$ drop below a certain tolerance. However, if the $u_j$ have converged to a two-point cycle (see \reffig{fig:residuals_pca}) this condition will never be triggered. Thus, it is important to characterize the possible limiting behaviors of the sequence of latent iterates. To do so, we study its {\em shape} using topological data analysis (TDA) \citep{skraba2012topological,perea2015sliding,tralie2018quasi}. Following prior works using TDA to analyze periodicity \cite{perea2015sliding,tralie2018quasi} we focus on the zeroth and first {\em persistent Betti numbers} --- denoted as $B_0$ and $B_1$ respectively --- of the point cloud $\{u_j\}_{j=1}^{K}\subset \mathbb{R}^n$. See \refappsec{app:TDA} for precise definitions. TDA examines the shape of the object formed by the union of small balls centered at each $u_j$. $B_0$ counts the number of connected components of this object, while $B_1$ counts the number of loops which encircle a hole \cite{munch2017user}. If the latent iterates have converged to a fixed point, then $B_0=1, B_1 = 0$, and this holds even if the convergence is only to within some small tolerance. If the latent iterates have converged to a limit cycle, $B_1 > 0$. By computing and interpreting $B_0$ and $B_1$ for a large number of input mazes, we identify three topologically distinct convergence behaviors. 

\section{Testing pre-trained models}
\label{sec:testing_pre-trained_models}

\begin{figure*}[t]
    \begin{tabular}{cccc}
        \multicolumn{2}{c}\DTNet & \multicolumn{2}{c}\PINet \\
        {\tiny\tt deadend\_start=True} & {\tiny\tt deadend\_start=False} &
        {\tiny\tt deadend\_start=True} & {\tiny\tt deadend\_start=False} \\
        \includegraphics[width=0.24\textwidth]{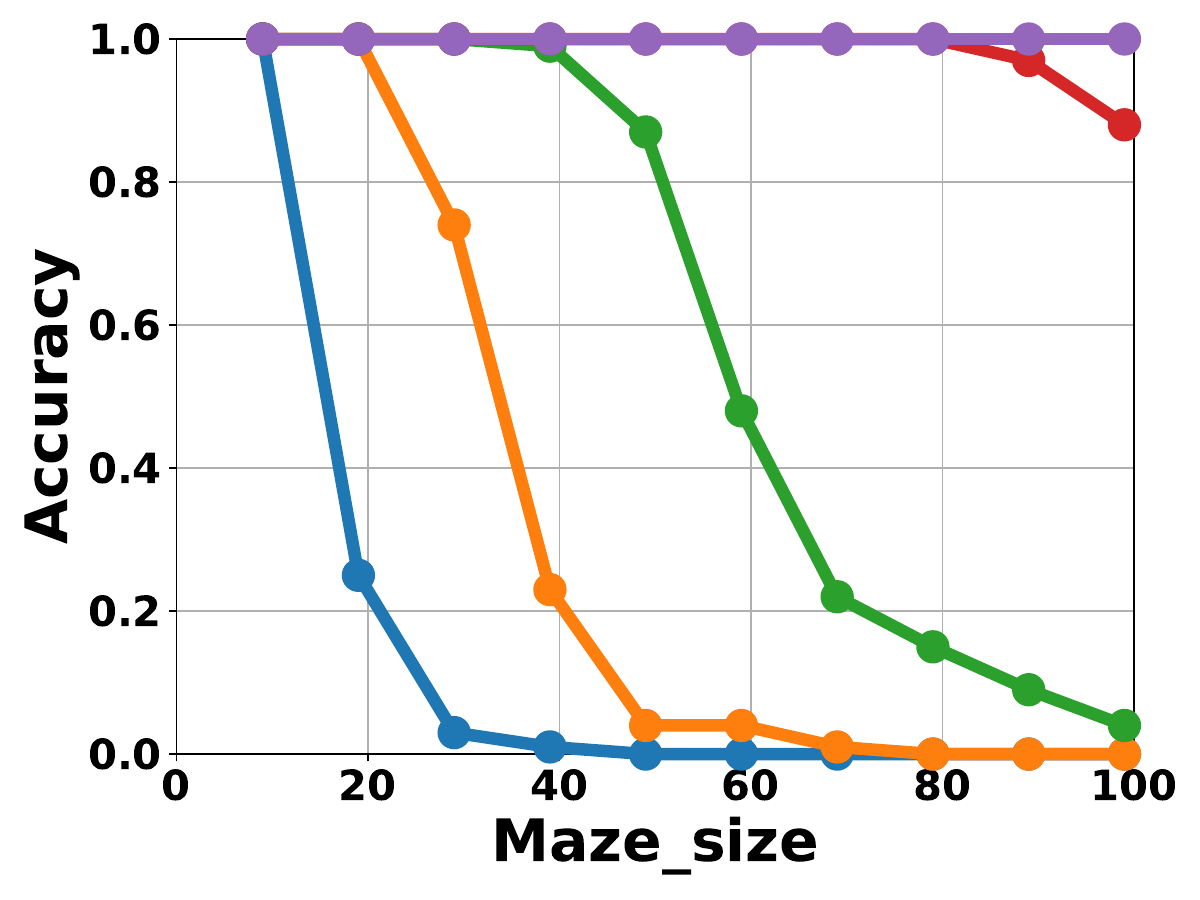} &
        \includegraphics[width=0.24\textwidth]{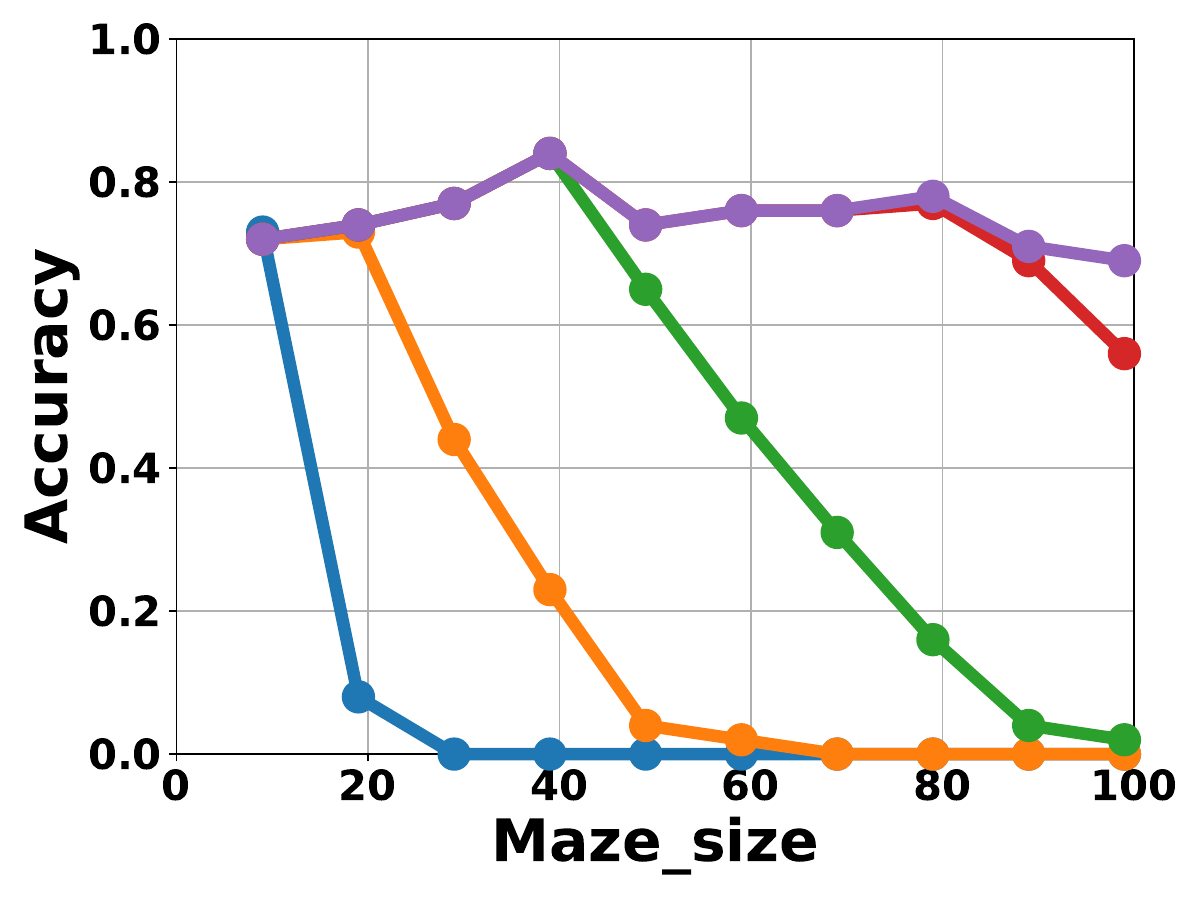} &
        \includegraphics[width=0.24\textwidth]{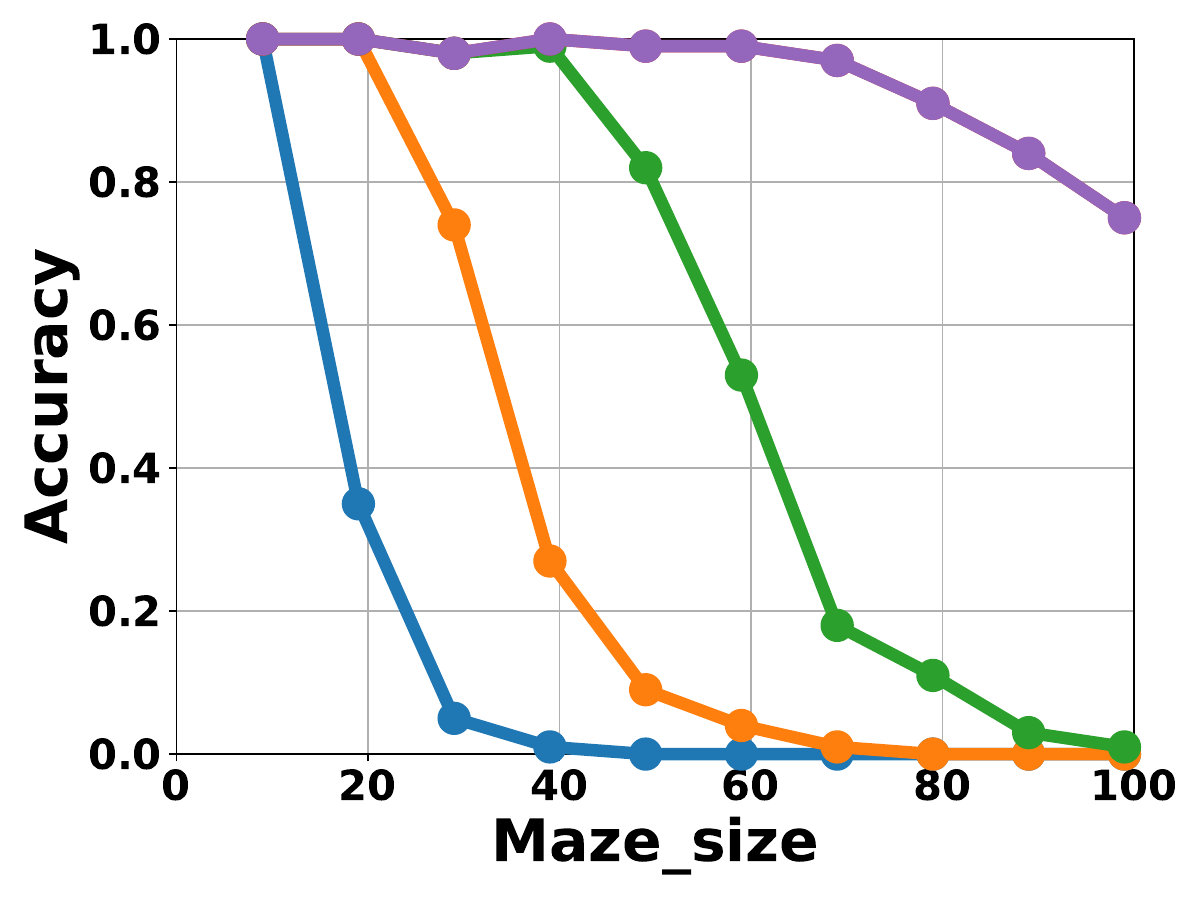} &
        \includegraphics[width=0.24\textwidth]{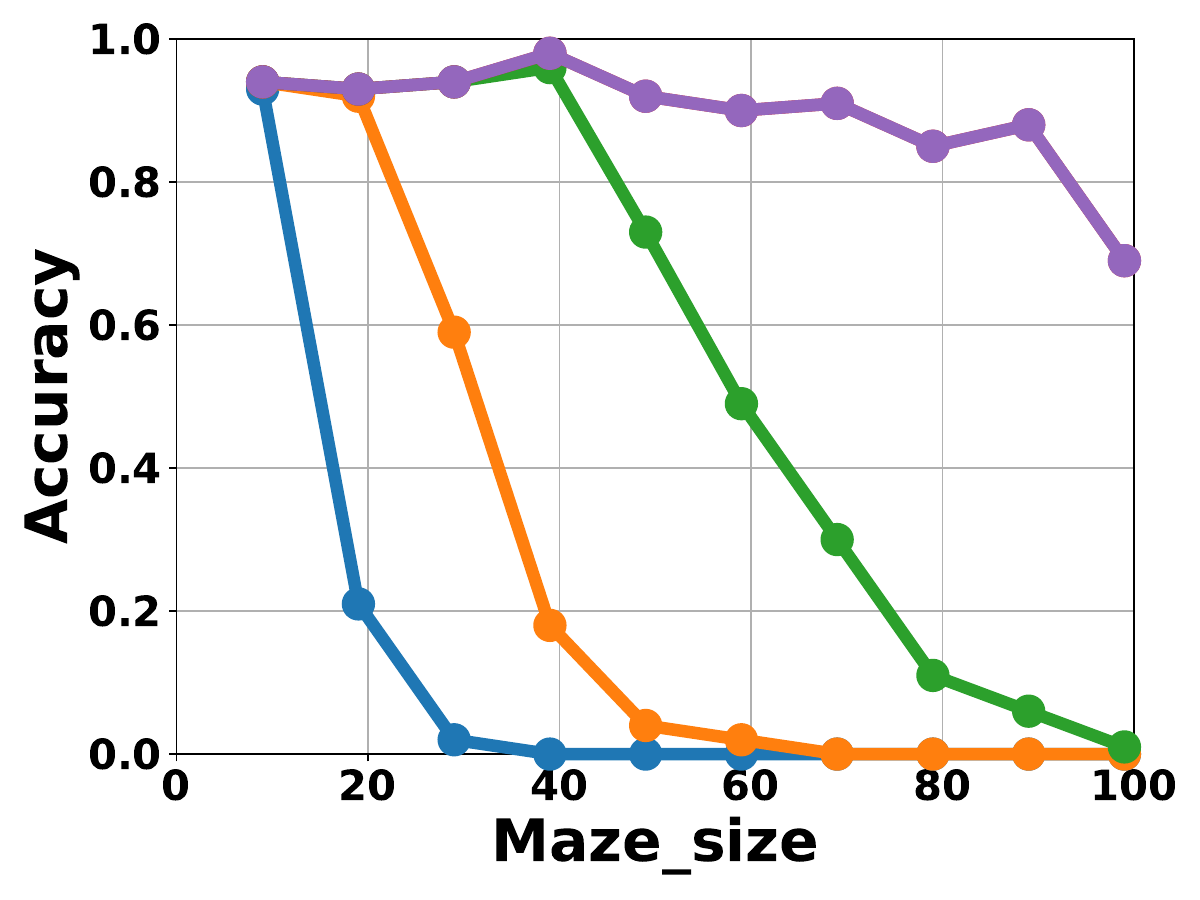} \\
    \end{tabular}
    \includegraphics[width=0.99\textwidth]{images/extrap_legend.pdf}
    \caption{
        {\bf Left:} \DTNet extrapolation accuracy (see \refsec{sec:maze-solving_task}) at various maze sizes, with {\tt deadend\_start=True} and {\tt deadend\_start=False}. {\bf Right:} Analogous results for \PINet. Both models extrapolate well to larger mazes given sufficient iterations. However, performance diminishes when the start position is allowed to have neighbors, regardless of the number of iterations. The sample size was limited to 100 mazes due to hours of compute time required for large $99 \times 99$ mazes.
    }
    \label{fig:extrap_deadend}
\end{figure*}

We study two trained maze-solving models from previous works: an RNN from \citet{bansal2022endtoend} which we call \DTNet, and an INN from \citet{anil2022path} which we call \PINet\footnote{\PINet stands for `Path-Independent' net, as path independence is a feature identified in \cite{anil2022path} as being strongly correlated with generalization. }. While both works propose multiple models, we focus on the most performant model from each. Both \DTNet and \PINet contain 0.78M parameters, facilitating fair comparison. \DTNet uses a progressive loss function to encourage improvements at each RNN layer. In this approach, the recurrent module is run for a random number of iterations, and the resulting output is used as the initial input for the RNN, while gradients from the initial iterations are discarded. The model is then trained to produce the solution after another random number of iterations. We refer the reader to~\cite[Section 3.2]{bansal2022endtoend} for additional details. For \PINet, path-independence is encouraged in two ways: (i) by using random initialization for half of the batch and zero initialization for the other half, and (ii) by varying the compute budgets/depths of the forward solver during training.

\subsection{Distribution shift}

Using {\tt maze-dataset} we explore the out-of-distribution behavior of \DTNet and \PINet. As shown in \reffig{fig:maze_extrap_visual} we vary the distribution along three dimensions: increasing maze size, setting {\tt deadend-start} to false, and increasing percolation.

\textbf{Maze size}. We first verify the extrapolation performance of \DTNet and \PINet with increasing maze size. For each maze size $n\times n$, where $n \in \{9, 19, 29, \ldots, 99\}$, we tested each model on 100 mazes. As expected, with sufficient iterations, both models achieve strong performance, confirming the results of \citep{bansal2022endtoend,anil2022path}. See the plots labeled {\tt deadend\_start=True} in \reffig{fig:extrap_deadend}. Both models achieve perfect accuracy on the $9 \times 9$ mazes of the training distribution. Furthermore, with 3,000 iterations, \DTNet achieves perfect accuracy and correctly solved all test mazes. \PINet achieved near perfect accuracy on smaller mazes, but performance diminished for mazes larger than $59 \times 59$. Importantly, running more iterations usually helps and never harms accuracy. Note that for \PINet the performance of the model after 1,000 iterations is identical to performance after 3,000 iterations at all tested maze sizes, suggesting that convergence occurred by 1,000 iterations.

\begin{figure}[t]
    \centering
    \includegraphics[width=0.8\linewidth]{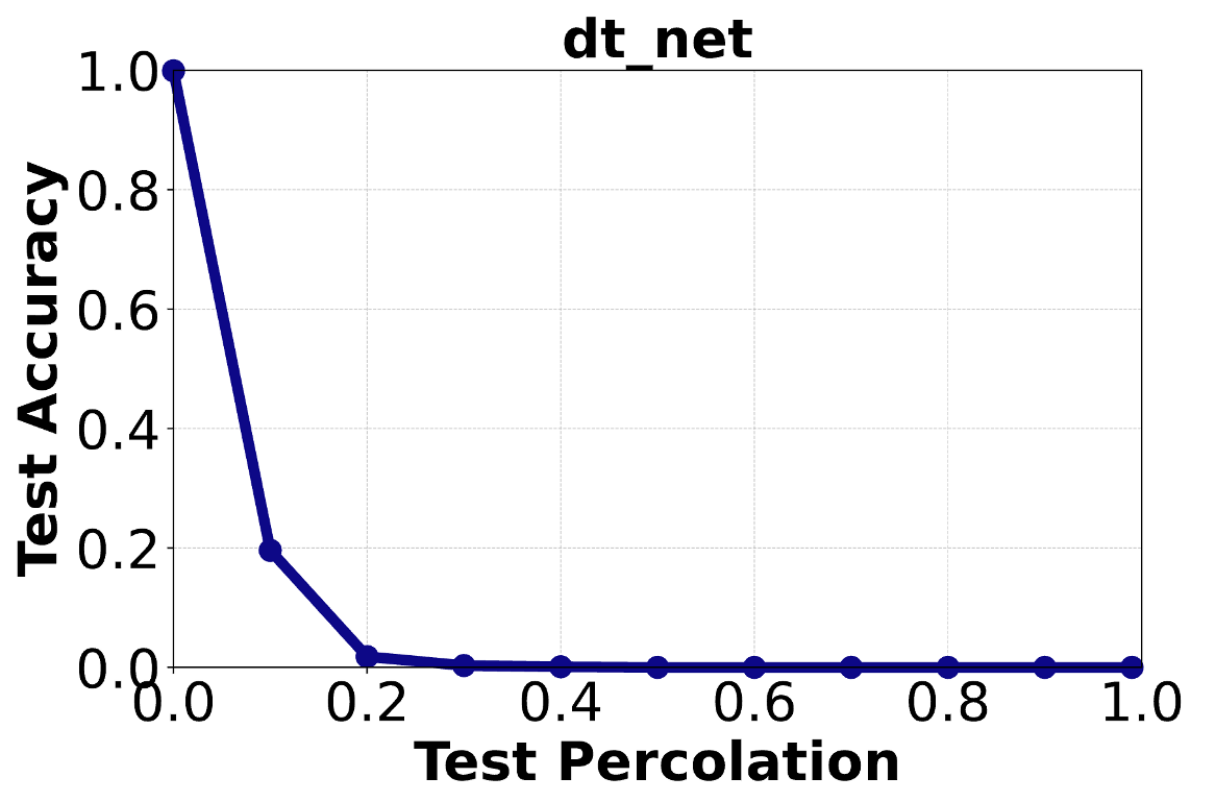}
    \caption{
        The accuracy of \DTNet rapidly diminishes when percolation increases above 0. \DTNet was iterated 30 times, and the resulting outputs do not change with additional iterations. The corresponding plot for \PINet is nearly identical.
    }
    \label{fig:extrap_percolation}
\end{figure}

\noindent \textbf{Deadend start}. Allowing the start position to have multiple neighbors, i.e., setting {\tt deadend\_start} to {\tt False}, represents a different out-of-distribution shift from the training dataset. This shift diminishes the performance of both models. See the plots labeled {\tt deadend\_start=False} in \reffig{fig:extrap_deadend}. With this shift, accuracy on $9 \times 9$ mazes drops from 1.00 to 0.72 for \DTNet and from 1.00 to 0.94 for \PINet. Interestingly, the fraction of failed predictions remains relatively stable as maze size is increased. There is no clear qualitative difference between mazes that were correctly and incorrectly solved by the models. Both models correctly solve some instances where the start position has multiple neighbors (see \refappfig{fig:extrap_no_deadend_split_neighbors}). However, we do observe that accuracy decreases monotonically as the degree of the start node increases. There is no discernible pattern to the models' failures; sometimes they are only one or two pixels away from the correct solution, sometimes they are missing large chunks of the correct path (see \refappsec{app:failed_model_predictions}).

\noindent \textbf{Percolation}. 
The final out-of-distribution shift we consider is increasing the percolation value of the test dataset from 0. We reiterate that when percolation equals 0, all mazes are acyclic. Increasing percolation adds cycles to the maze, introducing multiple solutions where there was previously exactly one. This shift significantly reduces accuracy, as shown in \reffig{fig:extrap_percolation}. Notably, increasing the number of iterations beyond 30 in this setting \emph{does not improve model performance}. 

\begin{figure}[t]
    \centering
    \includegraphics[width=0.9\linewidth]{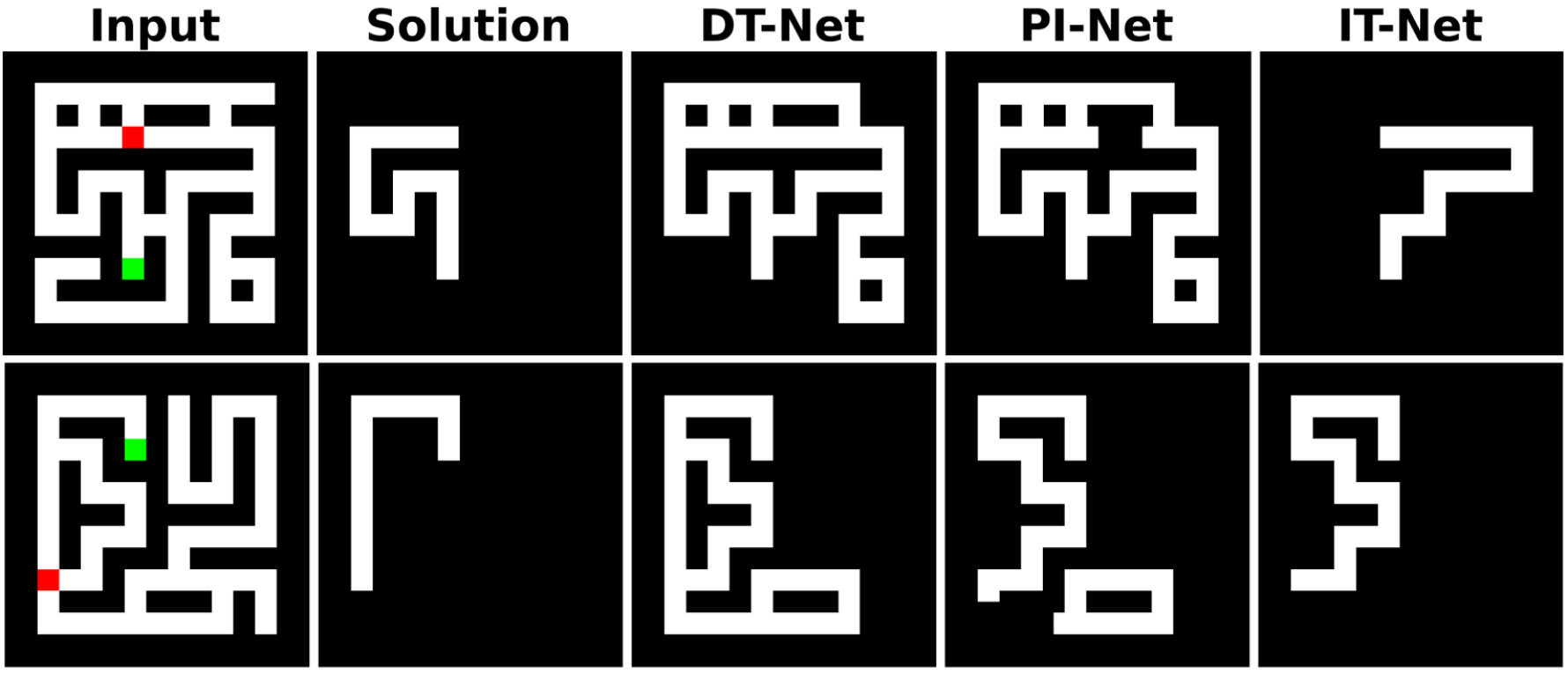}
    \caption{ Two examples of mazes with cycles. \DTNet fails like deadend-filling by retaining cycles from the input in its prediction. \PINet approximates deadend-filling but with some segments of white pixels removed. In the first example (top row), \ITNet predicts a minimal path (length 9) that is different from the solution (length 9); in the second example (bottom row), \ITNet predicts a valid non-minimal path (length 11) that is longer than the solution (length 8).}
    \label{fig:maze_cycles_example}
\end{figure}

\subsection{Latent dynamics}
\label{sec:latent_dynamics}

 In \citep[App. F, App. G]{anil2022path} evidence of instances $d$ where the latent iterates induced by $T_{\Theta_1}(\cdot, d)$ form a limit cycle, yet $\mathcal{N}_{\Theta}(d)$ is correct, is provided. Specifically, the residuals $r_j := \| u_{j+1} - u_j \|_2$, i.e. the distances between consecutive iterates, are considered. If $r_j = 0$ for all sufficiently large $j$ then the $u_j$ have converged to a fixed point. However, \citep{anil2022path} finds maze instances $d$ such that the residual sequence $\{r_j\}_{j=\tilde{K}}^{K}$ induced by a variant\footnote{Although not the variant we consider, see \refappsec{app:more_on_pi-net} for further discussion.} of \PINet is visually periodic.

\begin{figure}[t]
    \centering
    \includegraphics[width=0.35\linewidth]{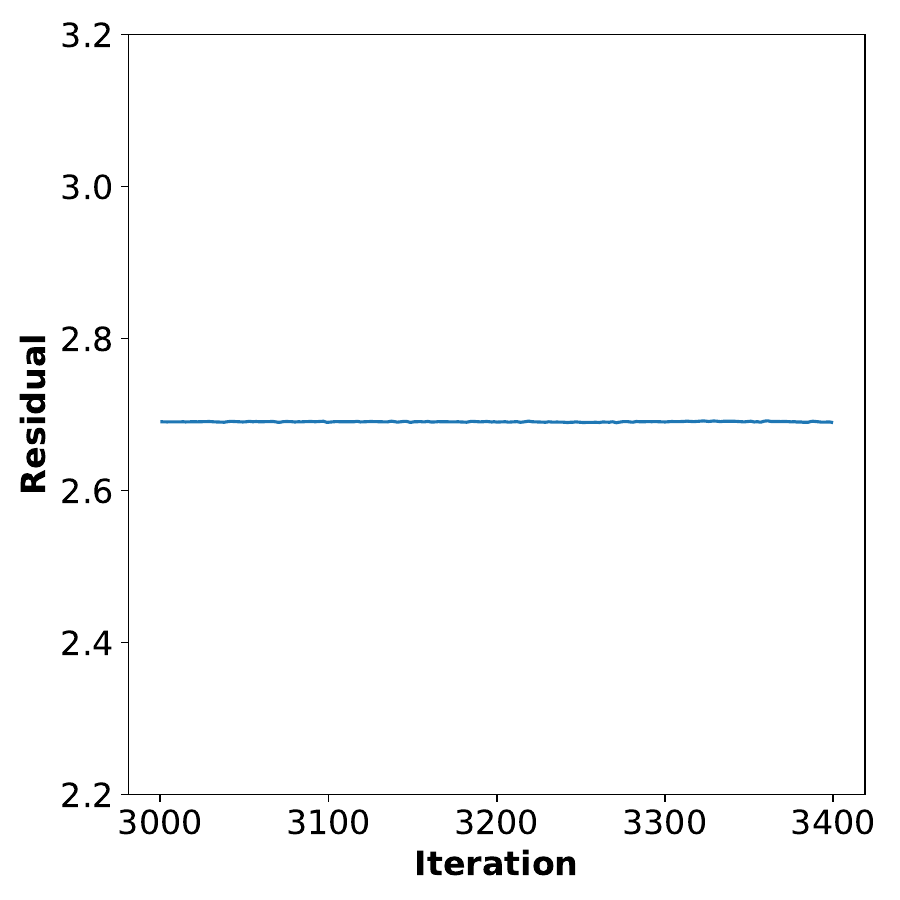}
    \includegraphics[width=0.4\linewidth]{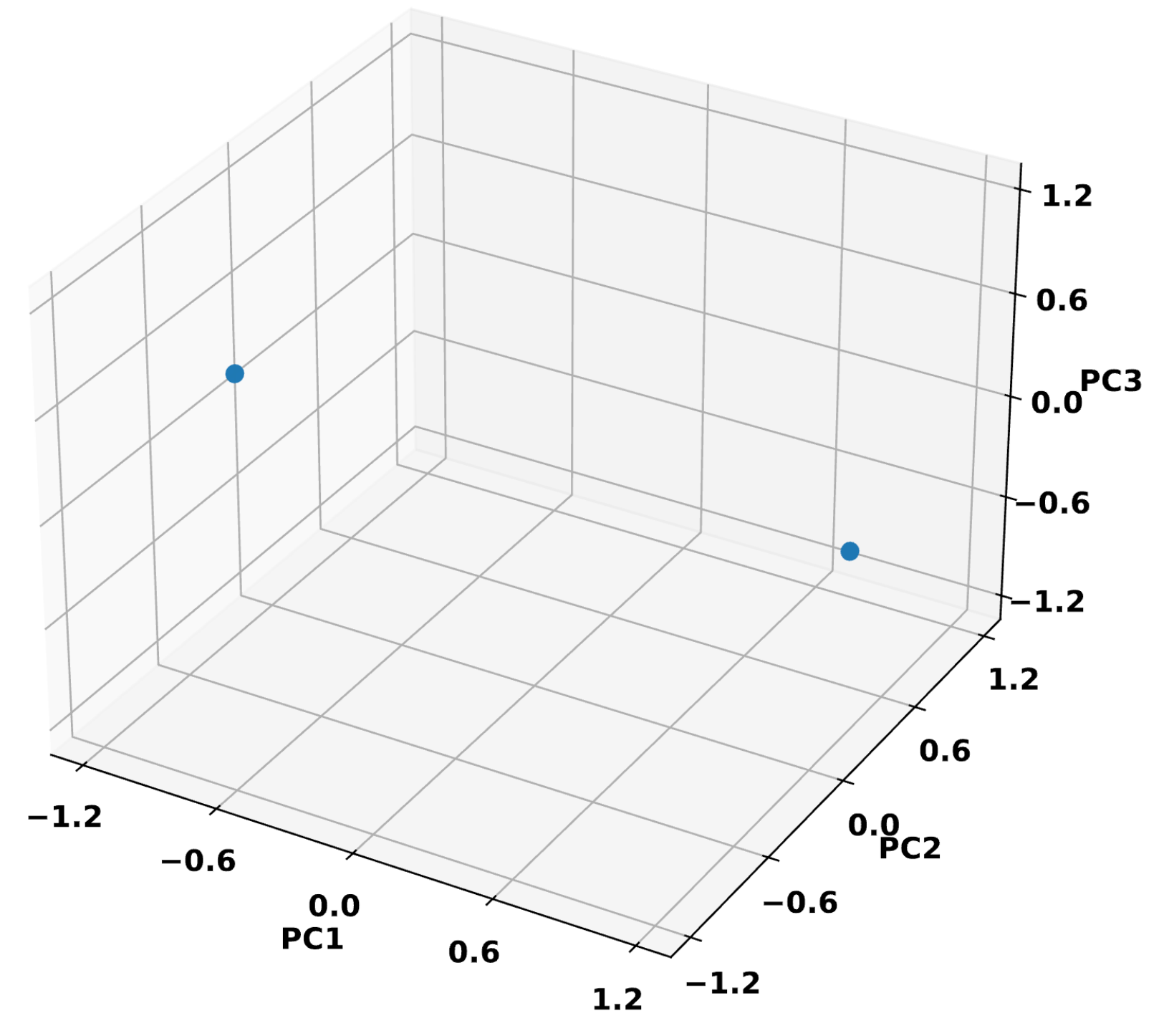}
    \includegraphics[width=0.35\linewidth]{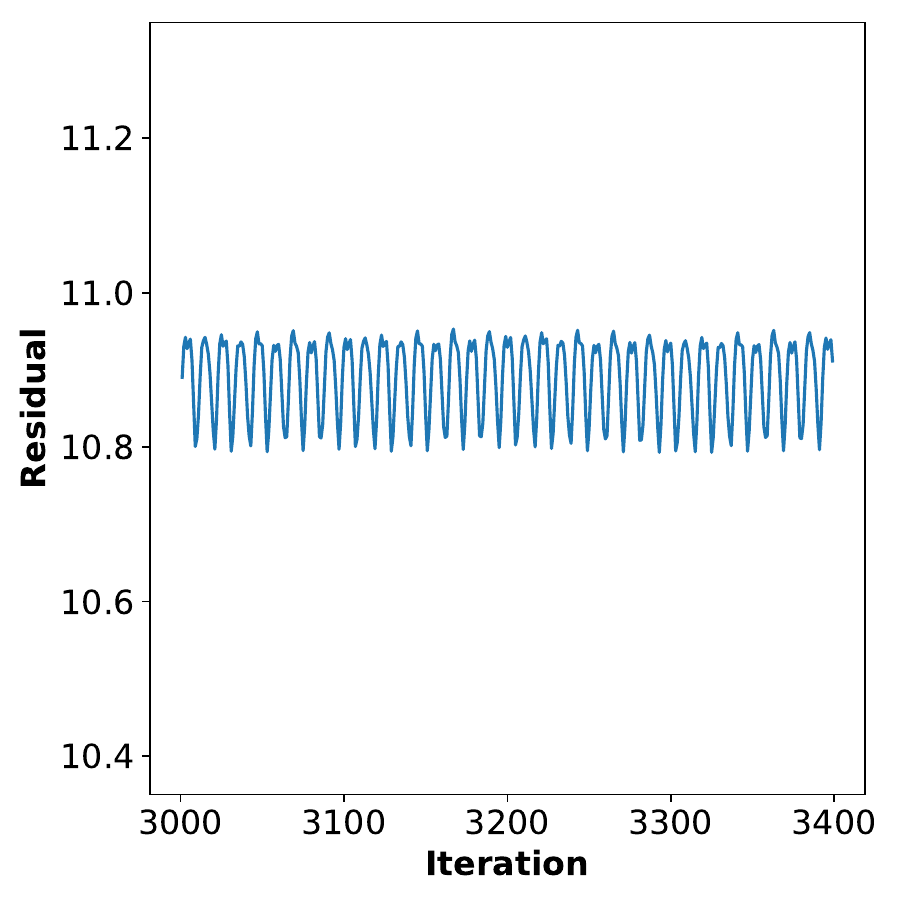}
    \includegraphics[width=0.4\linewidth]{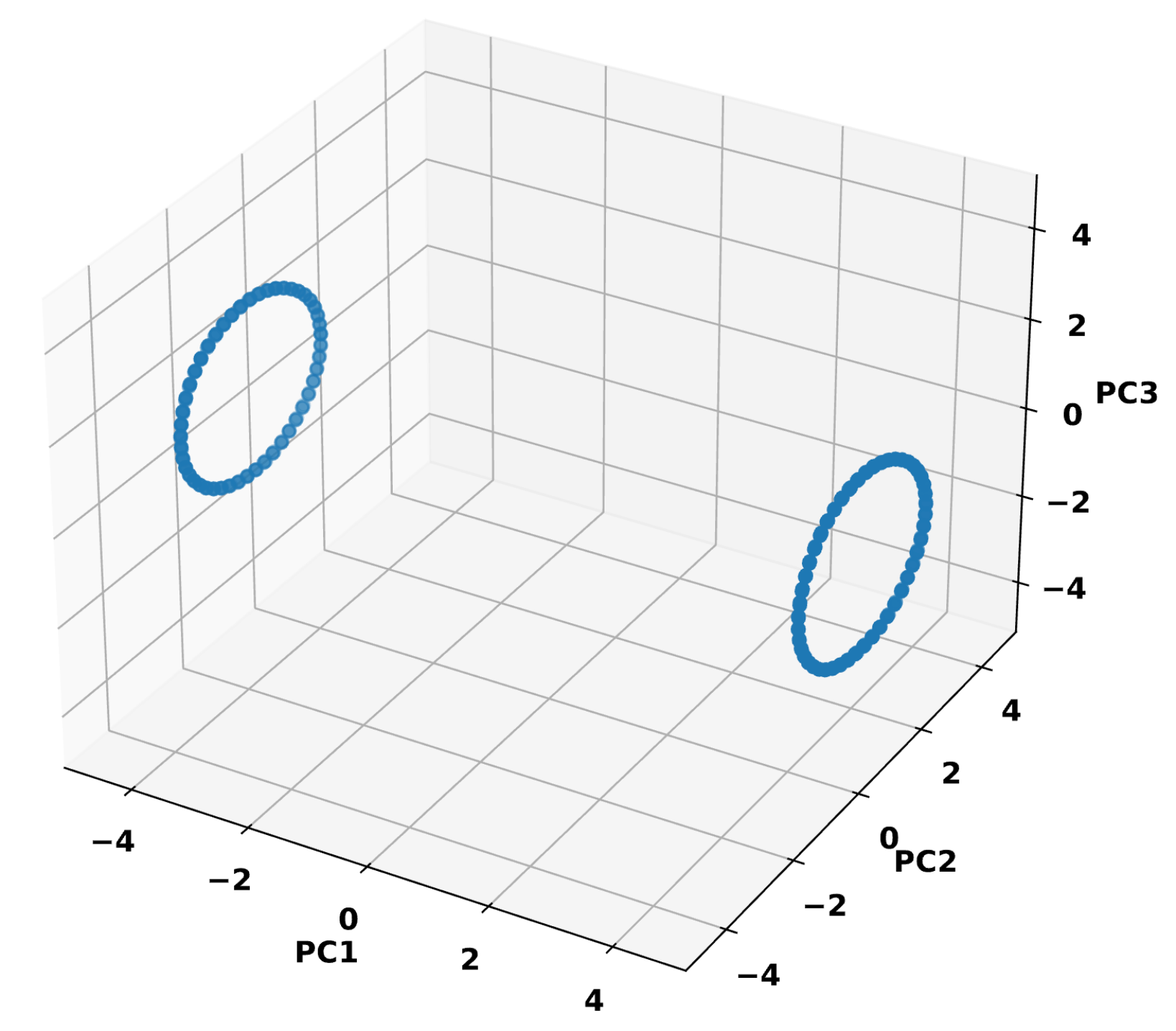}
    \caption{ Residual plots and corresponding PCA projections for two sequences of \DTNet latent iterates. The left two plots indicate oscillation between two points corresponding to $[B_0, B_1] = [2,0]$ for a $19 \times 19$ maze. The right two plots indicate oscillation between two loops corresponding to $[B_0, B_1] = [2,2]$ for a $69 \times 69$ maze. Both mazes were solved correctly.}
    \label{fig:residuals_pca}
\end{figure}

 We investigate the limiting behavior of \PINet and \DTNet further. For both models, we consider 100 $n\times n$ mazes where $n = 9,19,\ldots, 69$. We set $\texttt{percolation}=0$ and \texttt{deadend\_start=True}. We select a ``burn-in'' parameter $\tilde{K} < K$ and then consider latent iterates $\{u_j\}_{j=\tilde{K}}^{K}$. We set $\tilde{K} = 3,001$ and $K=3,400$. We observe periodicity of residuals for \DTNet (see \reffig{fig:residuals_pca}, third panel) and discover a novel asymptotic behavior of the residuals: convergence to a {\em nonzero value} (see \reffig{fig:residuals_pca}, panel 1). To explain both behaviors further, we introduce two tools.

\noindent \textbf{PCA}.
Projecting the high-dimensional latent iterates onto their first three principal components provides a glimpse into the underlying geometry of $\{u_j\}_{j=\tilde{K}}^{K}$ responsible for the observed residual sequences. We identify a sequence of latent iterates for \DTNet that oscillates between two points (see \reffig{fig:residuals_pca}, panel 2), yielding constant, non-zero, values of $r_j$ equal to the distance between these points. We call such limiting behavior a {\em two-point cycle}. We also identify a sequence of latent iterates for \DTNet that oscillates between two loops (see \reffig{fig:residuals_pca}, panel 4), yielding values of $r_j$ that oscillate around the (non-zero) distance between these loops. We call such limiting behavior a {\em two-loop cycle}. To the best of our knowledge, neither of these limiting behaviors has been observed previously in the latent dynamics of a logical extrapolator. 

\noindent \textbf{TDA}. To quantify the frequency with which these limiting behaviors occur, we use TDA (see \refsec{sec:TDA}). This is possible because these limiting behaviors are distinguishable using the zeroth ($B_0$) and first ($B_1$) persistent Betti numbers (see \refsec{sec:TDA} and \refappsec{app:TDA}).  Convergence to a point is associated to $[B_0,B_1] = [1,0]$, a two-point cycle is associated to $[B_0,B_1] = [2,0]$, while a two-loop cycle yields $[B_0,B_1] = [2,2]$. \reftab{tab:tda_results} summarizes the TDA results.

For \PINet, every latent sequence converges to a fixed-point. For \DTNet at every maze size the majority of latent sequences approach a two-point cycle, a minority approach a two-loop cycle, and a few approach some other geometry. We find no correlation between limiting behaviour and accuracy. \DTNet, when given $3000+$ iterations, achieves perfect accuracy (see \reffig{fig:extrap_deadend}, panel 1) even though, for a majority of larger mazes, it converges to a two-point or two-loop cycle.  On the other hand, although \PINet converges to a fixed point for all $69\times 69$ mazes we tested, it does not achieve perfect accuracy (see \reffig{fig:extrap_deadend}, panel 3).

\begin{table}[t]
    \centering
    \begin{tabular}{ll|rrrrrr}
        & & \multicolumn{6}{c}{$n$ for $n \times n$ maze} \\
        \multicolumn{1}{c}{\bf MODEL}  
        &$\mathbf{[B_0,B_1]}$
        &$\mathbf{9}$ &$\mathbf{19}$ &$\mathbf{29}$ &$\mathbf{39}$ &$\mathbf{49}$ &$\mathbf{59}$
        \\ \hline\hline 
                            &           &  & &   &  &  &             \\
        \DTNet        &$[1, 0]^*$ &15 &0 &0 &0 &0 &0          \\
                            &           &  & &   &  &  &            \\
                            &$[1,1]$ &6 &6 &0 &0 &0 &0        \\
                            &           &  & &   &  &  &            \\
                            &$[2,0]$ &79 &88 &95 &100 &98 &98        \\
                            &           &  & &   &  &  &          \\
                            &$[2,1]$ &0 &1 &0 &0 &0 &0        \\
                            &           &  & &   &  &  &          \\
                            &$[2,2]$ &0 &5 &5 &0 &2 &2        \\
        \hline
                            &           &  & &   &  &  &            \\
        \PINet        &$[1,0]^*$ &100 &100 &100 &100 &100 &100  \\
                \hline
                            &           &  & &   &  &  &            \\
        \ITNet        &$[1,0]^*$ &100 &100 &100 &100 &100 &100  \\
    \end{tabular}
    \caption{
        Betti number frequencies for \DTNet and \PINet. \PINet always exhibits fixed-point convergence ($[B_0, B_1] = [1,0]$) whereas \DTNet usually approach a two-point cycle ($[B_0, B_1] = [2,0]$) or sometimes a two-loop cycle ($[B_0, B_1] = [2,2]$). $^*$Sequence converged to within 0.01.
    }
    \label{tab:tda_results}
\end{table}

\begin{figure}
    \centering
    \includegraphics[width=0.4\linewidth]{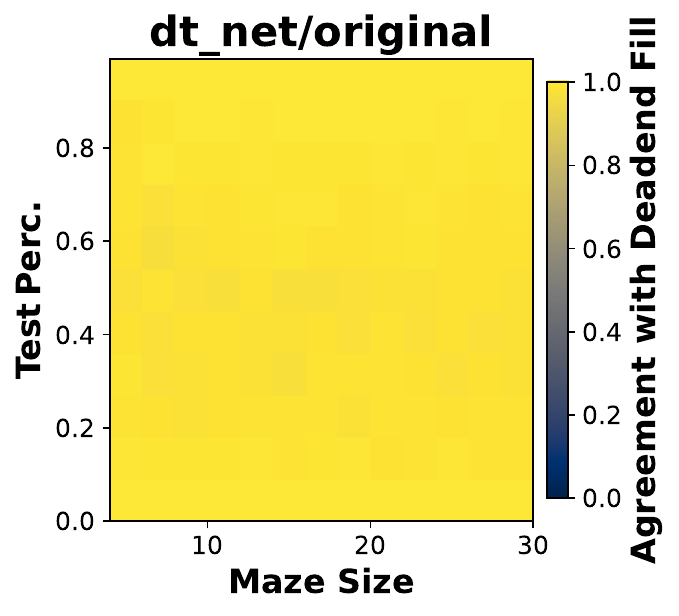}
    \includegraphics[width=0.4\linewidth]{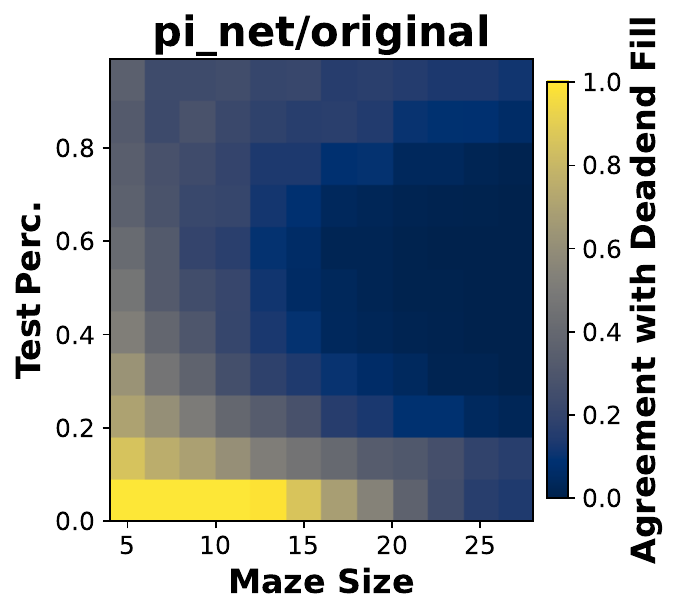} 
    \caption{
        Agreement of pretrained \DTNet and \PINet with deadend-fill algorithm, across 143K mazes with various size and percolation values. A value of 1.0 (yellow) means predictions perfectly coincide with deadend-filling algorithm, even when incorrect. \DTNet shows near perfect agreement, whereas \PINet predictions were often qualitatively similar but different by multiple pixels, resulting in low agreement.
    }
    \label{fig:agreement_heatmap}
\end{figure}
\section{The effect of diversifying training data}
\label{sec:diversifying_training_data}

\begin{figure*}[t]
    \centering
    \includegraphics[width=1.0\linewidth]{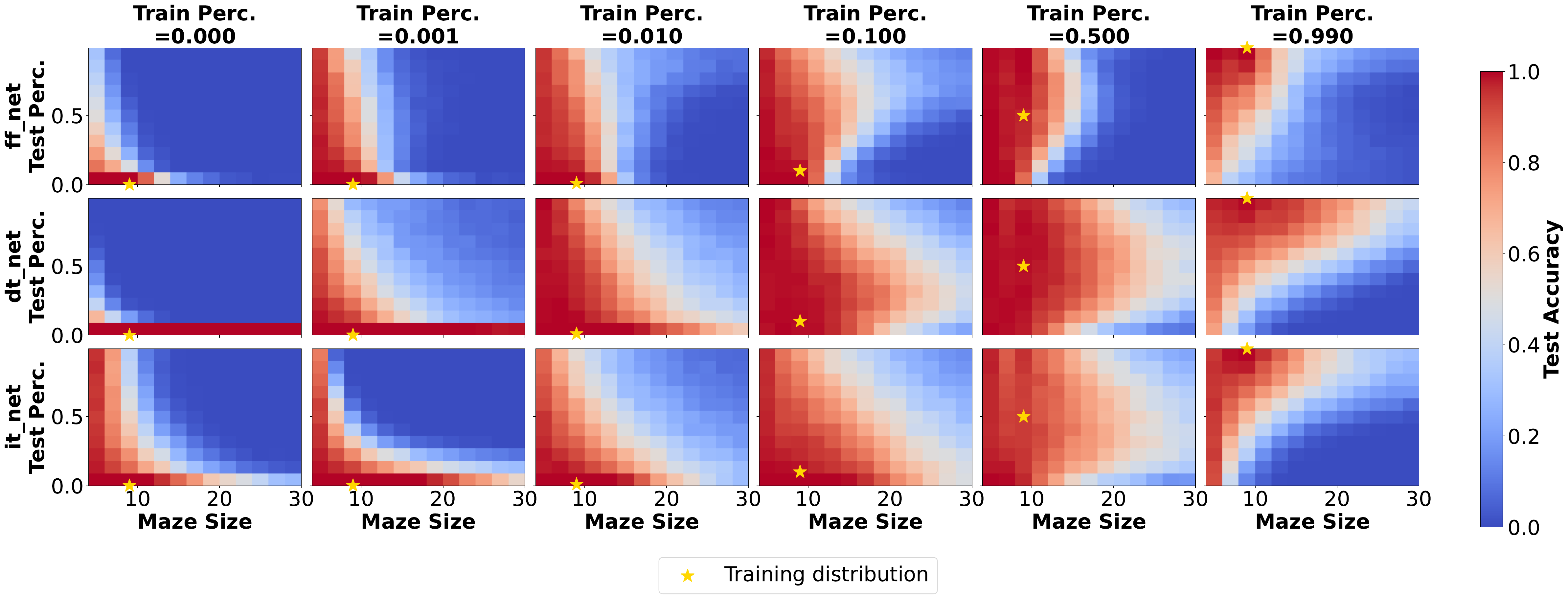}
    \caption{Heatmaps of test accuracy for \FFNet (top row), \DTNet (middle row) and \ITNet (bottom row) across various test maze sizes and test percolation values. A slight increase to training percolation (i.e. moving the gold star up) causes a dramatic increase in overall test accuracy. Each cell corresponds to 1,000 sampled test mazes.}
    \label{fig:heatmaps}
\end{figure*}

It is well known that increasing the size of the training dataset improves model generalization \cite{kaplan2020scalinglaws}. More recent works have also highlighted the importance of training dataset diversity and difficulty for generalization \cite{rolf2021subgroup-allocation, andreassen2021ood-robustness}. In \refsec{sec:testing_pre-trained_models} we identified the failure of \DTNet and \PINet to generalize to mazes with loops. We attempt to address this by training new models on diversified data containing some mazes with loops. We again use \DTNet but replace \PINet with a simpler implicit network we call \ITNet.\footnote{Although the pretrained model and associated code from \citep{anil2022path} is publicly available and suitable for inference, we encountered difficulties adapting the original implementation for training due to its complexity. To address this and facilitate reproducibility, we developed a simpler, more accessible training implementation.} We also train and test a model called \FFNet, a fully-convolutional feedforward network with a fixed depth of 30 layers, which serves as a baseline for generalization performance. Unlike \DTNet and \ITNet, \FFNet is incapable of test-time scaling. \FFNet is not weight-tied and so has roughly $10\times$ more weights than \ITNet. To diversify the training data we keep maze size fixed at $9\times 9$ while varying the percolation value. Then, we train randomly-initialized instances of \DTNet and \ITNet. We evaluate these models on test data with varying maze size and percolation value; see \reffig{fig:heatmaps}. For more training details, see \refappsec{app:training_details}.

\paragraph{Results} Increasing percolation shifts the training distribution closer to that of most test mazes. Thus the observed increase in overall test accuracy shown in \reffig{fig:heatmaps} and \refappfig{fig:overall_test_accuracy} is expected. We note the surprising jump in overall test accuracy---particularly for \DTNet---in response to a very small increase in percolation from 0 to 0.010. We interpret this switching behavior as a possible indication that diversification induces \DTNet to emulate a different class of algorithms capable of solving mazes with loops. However, this emulation is only approximate. \reffig{fig:heatmaps} shows that although the overall accuracy of both models increases as $p$ increases, their ability to extrapolate to larger mazes diminishes. We could not identify a clear pattern in $(n,p)$ values for which these models succeed, other than that performance degrades as the test distribution moves away from the training distribution. Consequently, it is hard to say with confidence that any of these models are behaving algorithmically. 

\DTNet and \ITNet have different architectures and were trained with different techniques. And yet \reffig{fig:heatmaps} and \refappfig{fig:overall_test_accuracy} show that both models generalize nearly equally well. This suggests that generalization is more sensitive to the diversity of the training dataset than the choice of model. Nonetheless, the results indicate specific models may be more suited for different extrapolation directions. \reffig{fig:heatmaps} shows \DTNet trained with $\texttt{percolation}=0$ perfectly solves larger mazes despite only seeing $9\times9$ mazes in training, and \ITNet trained with $\texttt{percolation}=0$ solves mazes with cycles better than the other models despite never seeing cycles in training.

\section{Have these networks learned an algorithm?}

In analyzing the failures of the pretrained \DTNet on mazes with cycles (see \reffig{fig:maze_cycles_example}) and \texttt{deadend\_start=True} we observed that the cycles remain part of the output. This is characteristic of the deadend-filling algorithm~\citep{hendrawan2020comparison} for maze solving. To study this further, we implemented deadend-filling, generated a large dataset (143k) of mazes with varying percolation and size, and compared the outputs of \DTNet to deadend-filling on these mazes. Our analysis reveals that \DTNet matches the output of deadend-filling --- succeeding when deadend-filling succeeds, yielding the same incorrect output when deadend-filling fails --- 98.8\% of the time (see \reffig{fig:agreement_heatmap}).  It appears that \DTNet has indeed learned to emulate\footnote{We use the word `emulate' to describe this form of agreement between a logical extrapolator and an algorithm. A stronger form of agreement is to match the iterates of the logical extrapolator to the iterates of the algorithm \citep{velivckovic2022clrs}. Some evidence that \DTNet and deadend-filling agree in this stronger sense is provided in \citep{schwarzschildalgorithm}.} deadend-filling, at least when \texttt{deadend\_start=True}.  

The situation with the pretrained \PINet is less clear. As shown in \reffig{fig:agreement_heatmap}, \PINet has low agreement with deadend-filling. However, a closer examination of \PINet predictions reveals that the disagreement with deadend-filling is often very minor. \PINet predictions are usually qualitatively very similar to deadend-filling. But for mazes that are large or contain many cycles, \PINet predictions match deadend-filling but with some segments of white pixels removed (see \reffig{fig:maze_cycles_example}), causing a low agreement. This slight difference with deadend-filling causes the disagreement shown in \reffig{fig:agreement_heatmap} and suggests that \PINet also has learned to approximate deadend-fill, but less successfully.

The models trained on diversified data showed superior overall accuracy (see \reffig{fig:heatmaps}) but worse logical extrapolation performance. Further, their predictions were less qualitatively consistent and thus difficult to identify with an algorithm.

\section{Limitations}
\label{sec:limitations}
One limitation of our work is that we limit ourselves to a single task. We focus on maze solving because it offers a variety of relatively intuitive distributional shifts (see \reffig{fig:maze_extrap_visual}). It would be interesting to apply our approach to other logical extrapolation tasks. A second, more philosophical weakness relates to what it means to `learn an algorithm', and what shifts of the test distribution should be considered reasonable.

\section{Discussion and Conclusion}
\label{sec:conclusion}
We present an in-depth look into logical extrapolation for a particular task: maze-solving. We confirm that by increasing the test-time compute budget (i.e., the number of iterations) certain models can solve much larger mazes than those in the training data, an impressive feat of OOD generalization. However, these models fail in interesting ways when the test distribution is shifted along other axes of difficulty. The observed failure modes indicate that logical extrapolators sometimes behave algorithmically, and we present evidence that \DTNet \cite{bansal2022endtoend} has indeed learned to emulate a simple algorithm for maze-solving, deadend-filling. As deadend-filling is known to fail on mazes with cycles, this also suggests that goal misgeneralization \citep{shah2022goal-misgeneralization, langosco2022goal-misgeneralization} may be the cause of the observed failures. Perhaps logical extrapolators are learning the simplest algorithm which fits the training data?

However, our experiments with training data diversification show that getting a neural network to learn an algorithm is a challenging task. Although adding mazes with cycles to the training data boosts model performance, the resulting models do not appear to behave algorithmically, and their ability to extrapolate to larger mazes is diminished. Moreover, we show that logical extrapolators can succeed even when their iterative part does not converge to a fixed point, and identify and quantify two exotic limiting behaviors. This complicates the view of logical extrapolators as fixed-point-finding routines \cite{bansal2022endtoend}. This also has consequences for training techniques which assume convergence to a fixed point \cite{fung2022jfb,geng2021training,gengattention,ramzi2022shine}.

To conclude, although there does appear to be something special about logical extrapolation, the usual caveats about OOD generalization still apply. Logical extrapolators may appear to behave algorithmically when the variation of the test distribution is tightly controlled, but respond in unexpected ways to other distribution shifts, much like other neural networks \cite{darestani2021measuring,gottschling2025troublesome,heckel2024deep}. Future work should focus on identifying conditions under which reliable algorithm learning is possible or reliable, and further exploring the relation of latent dynamics with logical extrapolation. 

\section*{Acknowledgments}
This work was supported by the National Science Foundation under Grant DMS‑2309810.

{\small
  \bibliography{references.bib}
}

\appendix

\section{Additional Maze Dataset Details}
\label{app:maze-dataset}

Both \DTNet and \PINet were trained on the same maze dataset \citep{schwarzschild2021datasets} containing $50,000$ mazes of size $9 \times 9$, meaning the mazes are subgraphs of the $5 \times 5$ lattice. Training mazes were generated via RDFS without percolation, and with the start position being constrained to being at a dead end (exactly $1$ neighbor). We mimic this training distribution and add out-of-distribution shifts using the {\tt maze-dataset} Python package \citep{ivanitskiy_maze-dataset_2023}.

We first note that for controlling maze size, \texttt{maze-dataset.MazeDatasetConfig}
takes a parameter \texttt{grid\_n}, which denotes the size of the lattice subgraph.
By contrast, the “easy to hard” \citep{schwarzschild2021datasets} dataset considers
an $n\times n$ maze in its raster representation.  To convert between these two
notions of maze size, we set
\[
  n = 2 \cdot \mathtt{grid\_n} - 1\,.
\]

We can modify the start position constraint in {\tt maze-dataset} by setting {\tt deadend\_start=False} in {\tt endpoint\_kwargs}. When {\tt False}, the start position is sampled uniformly at random from all valid nodes, while when {\tt True} the start position is samples uniformly at random from all valid nodes with degree one. Valid nodes are, by default, those not directly matching the end position or directly adjacent to it.

Randomized depth-first search (RDFS) is a standard algorithm for generating mazes, and produces mazes which are spanning trees of the underlying lattice, and thus do not contain cycles. For any acyclic graph with a spanning connected component, there is a unique (non-backtracking) path between any pair of points, and thus solutions are guaranteed to be unique. In our work, we select {\tt LatticeMazeGenerators.gen\_dfs\_percolation} as the {\tt maze\_ctor} parameter. This function takes an additional variable {\tt p} in {\tt maze\_ctor\_kwargs}, which controls the percolation parameter. This percolation parameter, denoted $p$ in our work, means that the final maze is the result of a logical {\tt OR} operation on the presence of all possible edges in the maze between an initial maze generated via RDFS and a maze generated via percolation, where each edge is set to exist with probability $p$. This is equivalent to first generating a maze with RDFS and then setting each wall to an edge with probability $p$. Since the initial RDFS maze is a spanning tree, adding any edge will cause the creation of a cycle, thus giving our desired out-of-distribution mazes.

\section{Topological Data Analysis}
\label{app:TDA}

Topological Data Analysis, or TDA, attempts to produce informative summaries of high dimensional data, typically thought of as point clouds\footnote{By using the term ``point cloud'' we are implying that the ordering of points does not matter.} $\mathcal{U} = \{u_1,\ldots, u_K\} \subset \mathbb{R}^n$, by adapting tools from algebraic topology. 

\subsection{Simplicial Complexes}
\label{sec:Simplicial_Complexes}

We are interested in TDA tools based on the idea of homology groups \cite[Chapter 2]{hatcher2002algebraic}. Homology groups can be computed algorithmically from a geometric object known as a {\em simplicial complex}, which we define below.
\begin{definition}
    \label{def:simplicial-complex}
    We collect definitions of several relevant concepts related to simplices. 
    \begin{enumerate}
        \item A $k$-simplex is the convex hull of any $k+1$ points in $\mathbb{R}^k$, 
        \begin{align}
            \sigma &:= \mathrm{Conv}\left\{u_1,\ldots, u_{k+1}\right\} \label{def:sigma}\\
                   & = \left\{\sum_{i=1}^{k+1}\alpha_iu_i: \ \alpha_i \geq 0 \text{ and } \sum_{i=1}^{k+1}\alpha_i = 1\right\} \nonumber
        \end{align}
    
        \item A face of a $k$-simplex $\sigma$ is a piece of the boundary of $\sigma$ which is itself a simplex. That is, $\tau$ is a face of $\sigma$ defined in \eqref{def:sigma} if
        \begin{equation*}
            \tau = \mathrm{Conv}\left\{u_{i_1},\ldots, u_{i_{\ell+1}}\right\}
        \end{equation*}
        \item A simplicial complex $\mathcal{S}$ is a set of simplices, of multiple dimensions, satisfying the following properties
        \begin{enumerate}
            \item For all $\sigma \in \mathcal{S}$, all faces of $\sigma$ are also in $\mathcal{S}$.
            \item If any two $\sigma_1,\sigma_2 \in \mathcal{S}$ have non-empty intersection, then $\sigma_1\cap\sigma_2$ is a face of both $\sigma_1$ and $\sigma_2$, and consequently $\sigma_1\cap\sigma_2 \in \mathcal{S}$
        \end{enumerate}
    \end{enumerate}
\end{definition}

The process of computing homology groups from a simplicial complex is beyond the scope of this paper. We refer the reader to \cite[Chapter 2]{hatcher2002algebraic} for further details. It is also possible to define homology groups for {\em abstract simplicial complexes}, $\mathcal{R}$, for which a $k$-simplex $\sigma \in \mathcal{R}$ is not literally a convex hull, but merely a list of points:
\begin{equation}
    \sigma = \{u_1,\ldots, u_{k+1}\}.
\end{equation}

In this case, a face is just a subset of $\sigma$:
\begin{equation}
    \tau = \left\{ u_{i_1},\ldots, u_{i_{\ell+1}} \right\}.
\end{equation}

Note that condition 3 (b) of \refappdef{def:simplicial-complex} is now vacuously true.

\subsection{Homology Groups}
Although we have not stated exactly how homology groups are defined, in this section we discuss a few of their properties. Fix a (possibly abstract) simplicial complex $\mathcal{S}$. We shall work with homology groups with coefficients in the field $\mathbb{Z}_2 := \mathbb{Z}/2\mathbb{Z}$, hence all homology groups will be vector spaces over $\mathbb{Z}_2$. We will focus on the zeroth and first homology groups, denoted $H_0(\mathcal{S})$ and $H_1(\mathcal{S})$ respectively. Elements in $H_0(\mathcal{S})$ are equivalence classes of points, where two points are equivalent if they are in the same path component of $\mathcal{S}$. Elements in $H_1(\mathcal{S})$ are equivalence classes of closed loops, where two loops are equivalent if they ``encircle the same hole'' \citep{munch2017user}. Consequently, the dimension of $H_0(\mathcal{S})$ (the zeroth {\em Betti number}, $B_0(\mathcal{S}$) counts the number of path connected components of $\mathcal{S}$, while the the dimension of $H_1(\mathcal{S})$ (the first {\em Betti number}, $B_1(\mathcal{S}$) counts the number of distinct holes in $\mathcal{S}$.

\subsection{The Rips Complex}
\label{sec:Rips_Complex}
Given the above, in order to associate Betti numbers to a point cloud we first need to define an appropriate simplicial complex.
\begin{definition}[The Vietoris-Rips complex]
\label{def:Rips_complex}
    Fix a point cloud $\mathcal{U} = \{u_1,\ldots, u_K\} \subset \mathbb{R}^n$, and for simplicity assume that $K < n$. Select a distance parameter $\epsilon \geq 0$. We define the simplicial complex $\mathcal{S}_{\epsilon}$, known as the Vietoris-Rips, or simply Rips, complex to contain all simplices 
    \begin{equation}
        \sigma = \mathrm{Conv}\left\{u_{i_1},\ldots, u_{i_{\ell+1}}\right\},
    \end{equation}
    satisfying the condition:
    \begin{equation}
        \max_{1\leq m < n \leq \ell+1} \|u_{i_m} - u_{i_n}\|_2 \leq \epsilon.
    \end{equation}
\end{definition}

In words, $\mathcal{S}_{\epsilon}$ contains all simplices on $\mathcal{U}$ with a diameter less than $\epsilon$. We note that $\mathcal{S}_0 = \mathcal{U}$ and hence $B_0(\mathcal{S}_0) = |\mathcal{U}|$, as every point in $\mathcal{U}$ is its own connected component., while  $B_1(\mathcal{S}_0) = 0$. On the other end of the scale, when $\epsilon > \mathrm{diam}(\mathcal{U})$, where 
\begin{equation}
   \mathrm{diam}(\mathcal{U}) =  \max_{1\leq i < j \leq N} \|u_{i} - u_{j}\|_2,
\end{equation}

the full-dimensional simplex
\begin{equation}
    \sigma = \mathrm{Conv}\left\{u_{1},\ldots, u_{K}\right\}
\end{equation}

is contained in $\mathcal{S}_{\epsilon}$, and thus $\mathcal{S}_{\epsilon}$ has one connected component and no loops: $B_0(\mathcal{S}_{\epsilon}) = 1$, $B_1(\mathcal{S}_{\epsilon}) = 0$. Consequently, the important topological features of $\mathcal{U}$ are detected by the Rips complex for $\epsilon$ values in $(0,\mathrm{diam}(\mathcal{U}))$.

The condition $K < n$ in \refappdef{def:Rips_complex} may be removed, in which case $\mathcal{S}_{\epsilon}$ is defined as an abstract simplicial complex. This distinction is not relevant for our work. 

\subsection{Persistent Betti Numbers}
\label{sec:Betti_Numbers}

Which value of $\epsilon$ should one choose? As discussed in \citet{munch2017user}, the trick is not to select a particular value of $\epsilon$ but rather focus on features (concretely: equivalence classes in $H_0(\mathcal{S}_{\epsilon})$ and $H_1(\mathcal{S}_{\epsilon})$) which {\em persist} for large ranges of $\epsilon$. More specifically, we define $\epsilon_b$, the {\em birth time}, to be the value of $\epsilon$ at which a particular equivalence class is first detected in $\mathcal{S}_{\epsilon}$. The death time, $\epsilon_{d}$, is the largest value of $\epsilon$ for which a particular equivalence class is detected in $\mathcal{S}_{\epsilon}$. Fixing a threshold $\text{\tt thresh}$, we say an equivalence class is {\em persistent} if $\epsilon_{d} - \epsilon_b > \text{\tt thresh}$. We define the persistent zeroth (respectively first) Betti number $B_0$ (respectively $B_1$) to be the number of distinct equivalence classes appearing in $H_0(\mathcal{S}_{\epsilon})$ (respectively $H_1(\mathcal{S}_{\epsilon})$) which satisfy $\epsilon_{d} - \epsilon_b > \text{\tt thresh}$. 

\subsection{Sliding Window Embedding}
Finally, we note that \citet{perea2015sliding,tralie2018quasi} propose a more sophisticated method for detecting periodicity using the {\em sliding window embedding}, also known as the {\em delay embedding}:
\begin{equation}
    \left\{u_j\right\}_{j=1}^{K} \mapsto \left\{ SW_{d,\tau}(u_j) := \begin{bmatrix} u_j \\ u_{j+ \tau} \\ \vdots \\ u_{j+d\tau} \end{bmatrix}\right\}_{j=1}^{K-d\tau}, 
\end{equation}
where $\tau$ (the delay) and $d$ (the window size) are user-specified parameters. Then, persistent Betti numbers are computed for $\{SW_{d,\tau}(u_j)\}_{j=1}^{K-d\tau}$ instead. This construction is motivated by Taken's theorem \cite{takens2006detecting} which, informally speaking, states that the dynamics of $\left\{u_j\right\}_{j=1}^{K}$ can be recovered completely from its sliding window embedding, for sufficiently large $d$. Note this comes at a price: the increase in dimension means an increase in computational cost.

In preliminary experiments we found that the persistent Betti numbers for $\{SW_{d,\tau}(u_j)\}_{j=1}^{K-d\tau}$ did not reveal anything that could not already be inferred from the persistent Betti numbers of $\{u_j\}_{j=1}^{K}$. Thus, we chose not to work with the sliding window embedding.

\section{Additional Experimental Details}
\label{app:experimental_details}

\subsection{Models Summary}
\label{app:models_summary}
\begin{table*}
  \small
  \centering
  \caption{{\small Summary of models used in experiments.}}
  \label{tab:model_summary}
  \begin{tabular}{llcc l}
    \hline
    \textbf{Model} & \textbf{Description} & \textbf{Params (M)} & \textbf{Size (MB)} & \textbf{Trained by} \\
    \hline
    \FFNet & fully‐convolutional feedforward            & 8.89 & 33.91 & Us \\
    \PINet & INN            & 0.78 & 2.99  & Anil et al.\ \cite{anil2022path} \\
    \DTNet & weight‐tied, input‐injected RNN             & 0.78 & 2.99  & Bansal et al.\ \cite{bansal2022endtoend} and Us \\
    \ITNet & INN variant of \DTNet & 1.37 & 5.24  & Us \\
    \hline
  \end{tabular}
\end{table*}

\subsection{Training Details}
\label{app:training_details}
For each model type, many variations of hyperparameters such as learning rate, gradient clipping, contractivity, tolerance, warmup, Jacobian-free backpropagation \citep{fung2022jfb}, and others were explored, and the best-performing models were selected to serve as representative examples. The training runs in \refsec{sec:diversifying_training_data} all utilized a 90/10 training/validation split of 100k mazes over 100 epochs, learning rate of $10^{-4}$, {\tt AdamW} optimizer, and {\tt ReduceLROnPlateau} learning rate scheduler with patience 10 and reduction factor 0.1. \DTNet was trained with gradient clip of 1.0 and progressive loss from \cite{bansal2022endtoend} with $\alpha = 0.01$. \ITNet was trained without gradient clipping, maximum iterations 100 with convergence tolerance 0.1, and using Jacobian-free backpropagation \cite{fung2022jfb}. See the code provided in the supplemental material for full details.

\subsection{Additional Details on \PINet}
\label{app:more_on_pi-net}
During our experiments we determined that there was another model parameter, beyond the number of iterations, that had a strong impact on the accuracy of \PINet. Specifically, there is a {\tt threshold} parameter within the forward solver, Broyden's method, that controls the maximum rank of the inverse Jacobian approximation. Based on code included in the supplementary material for \citet{anil2022path}, it appears the {\tt threshold} parameter was originally set at 40. However, with this setting \PINet performed very poorly; it failed on all mazes of size $49 \times 49$. Increasing {\tt threshold} increased the accuracy of \PINet, but also increases memory costs because it requires storing a number of high-dimensional latent iterates equal to {\tt threshold}. For our experiments, we used {\tt threshold = 1,000} in order to achieve strong accuracy without unreasonable memory requirements.

\subsection{Computational Resources}
\label{app:computational_resources}
The experiments in this study were performed on a high-performance workstation with the following specifications:
\begin{itemize}
    \item \textbf{CPU:} AMD Ryzen Threadripper PRO 3955WX (16 cores, 32 threads)
    \item \textbf{GPU:} 3x NVIDIA RTX A6000 (48 GiB VRAM)
    \begin{itemize}
        \item CUDA Version: 12.5, Driver Version: 555.42.06
    \end{itemize}
    \item \textbf{Memory:} 251 GiB RAM
    \item \textbf{Operating System:} Ubuntu 20.04.6 LTS (x86\_64 architecture)
\end{itemize}
as well as a node of a High Performance Computing platform with the following specifications:
\begin{itemize}
    \item \textbf{CPU:} AMD EPYC 9534 (64 cores, 128 threads)
    \item \textbf{GPU:} 4x NVIDIA RTX A6000 (48 GiB VRAM)
    \item \textbf{Memory:} 768 GiB RAM
    \item \textbf{Operating System:} Ubuntu 20.04.6 LTS Server.
\end{itemize}

Testing of two pre-trained models in \refsec{sec:testing_pre-trained_models} took about 40 GPU hrs per model or 80 GPU hours total. Training of the models in \refsec{sec:diversifying_training_data} took about 10 GPU hours per model and training distribution pair, or about 180 GPU hours total. The full research project required much more compute due to hundreds of GPU hours spent on hyperparameter selection. The final hyperparameters used in experiments are provided in the supplemental material.

\section{Additional Experimental Results}
\subsection{Failed Model Predictions}
\label{app:failed_model_predictions}

\begin{figure}
    \begin{tabular}{cc}
         \DTNet 
         & \PINet
         \\
         \includegraphics[width=0.49\linewidth]{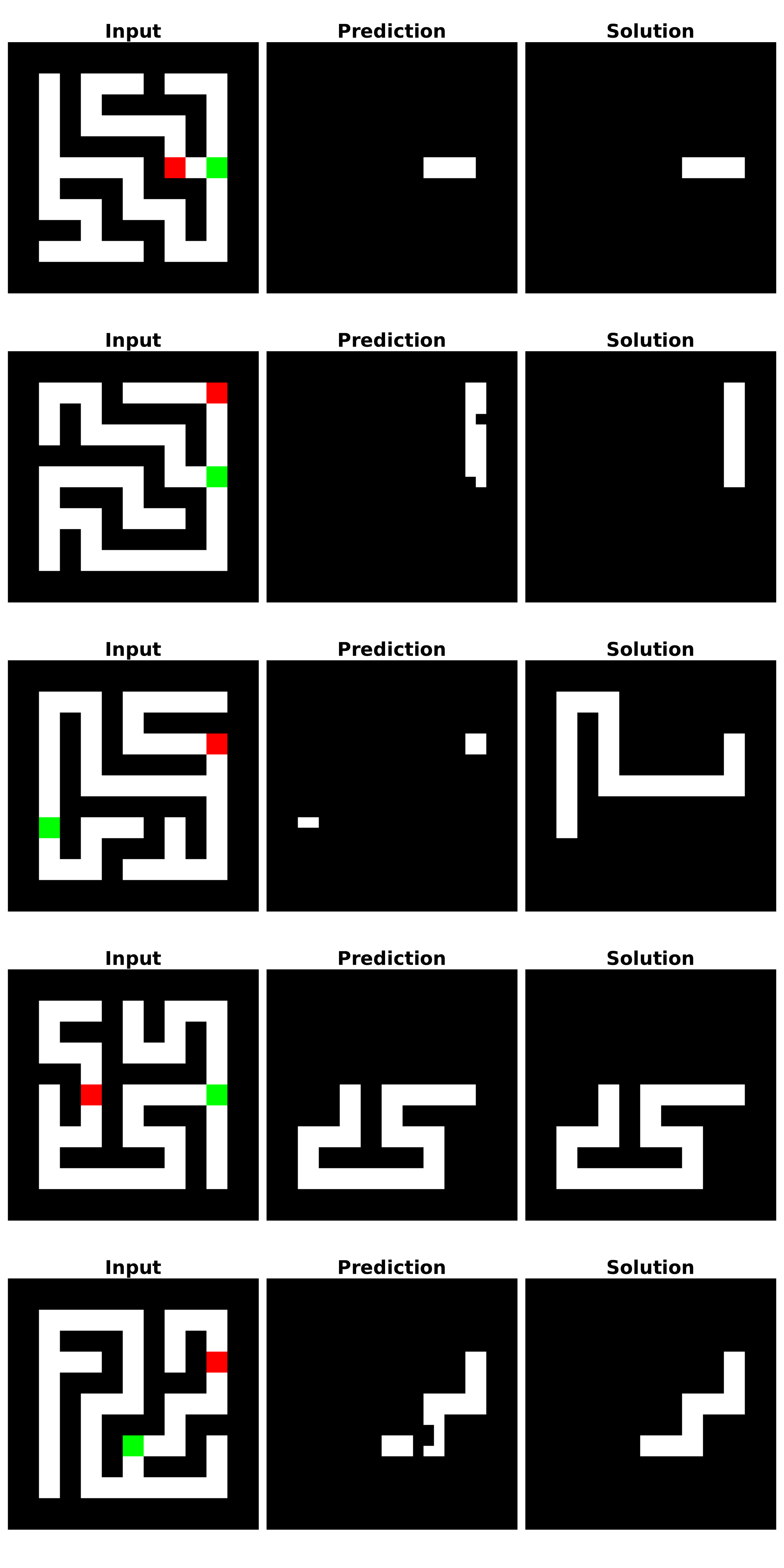}
         & \includegraphics[width=0.49\linewidth]{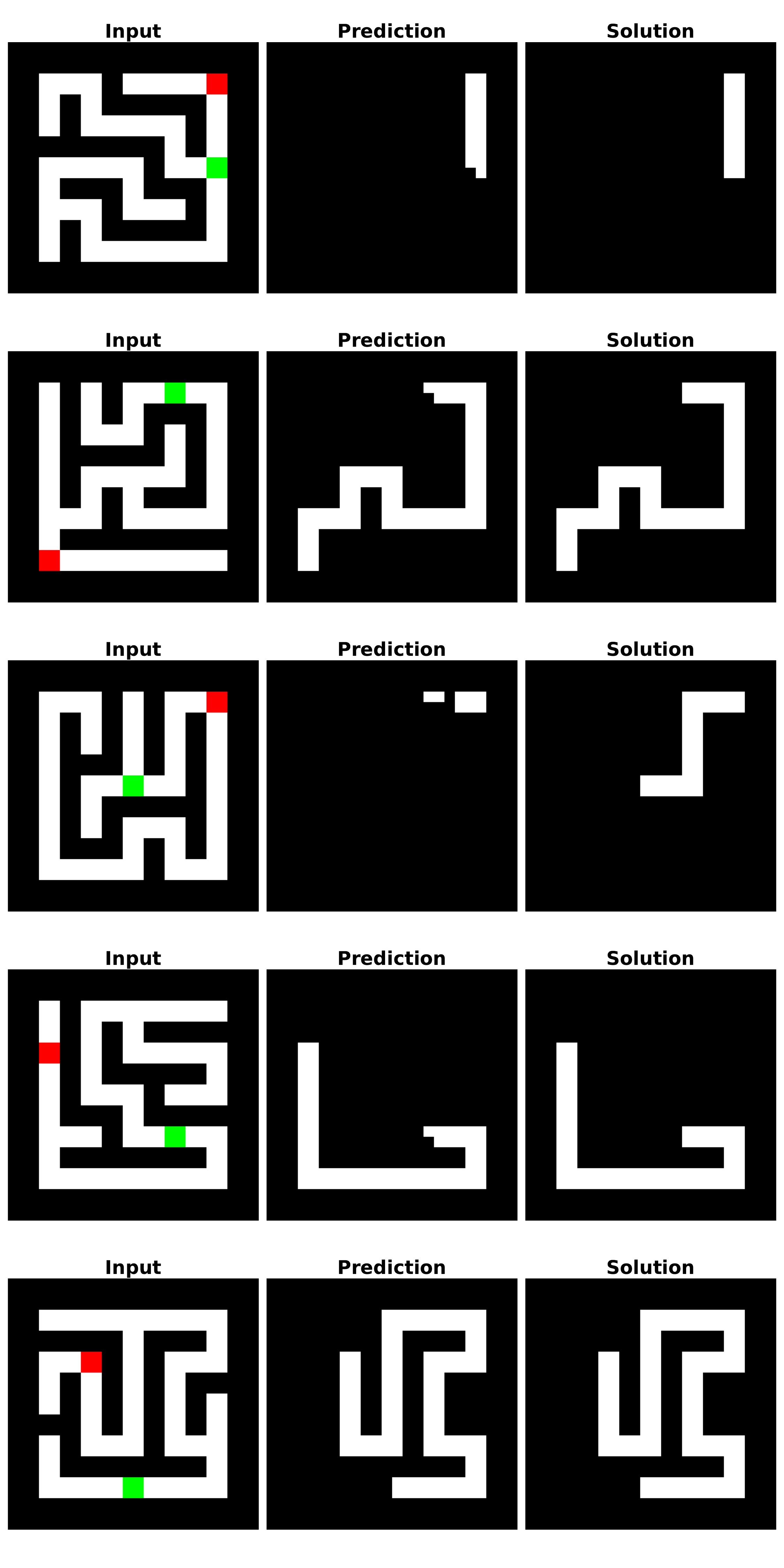}
         \\
    \end{tabular}
    \caption{{\small Examples of $9 \times 9$ mazes with {\tt deadend\_start=False} predictions from \DTNet (left) and \PINet (right), some of which they fail to solve. Note that mistakes are often in the start position cell or cells immediately adjacent to it.}}
\end{figure}

\subsection{Additional Experimental Results}
\label{sec:additional_test_accuracy_plots}

\begin{figure}
    \includegraphics[width=1.0\linewidth]{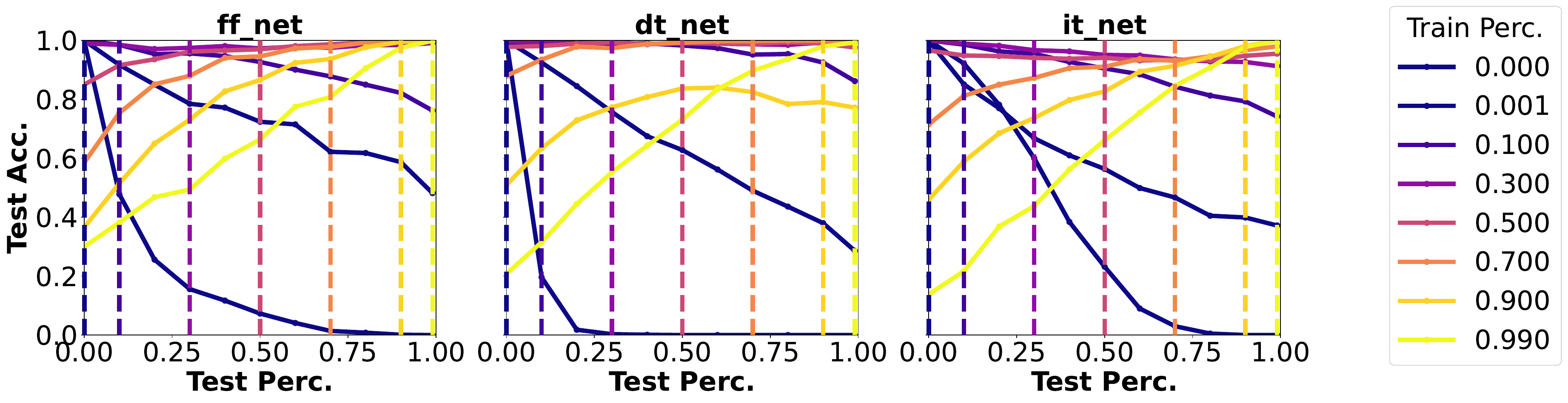}
    \caption{{\small Alternative visualizations of test accuracy for \FFNet, \DTNet, and \ITNet across different test percolation values at maze size 9 (the training maze size). Each accuracy value is computed from a sample of 1000 test mazes. The dashed vertical lines illustrate the training percolation value of the model with the matching color.}}
    \label{fig:additional_test_acc}
\end{figure}

\begin{figure}
    \centering
    \begin{tabular}{cc}
        \DTNet & \PINet \\
        \includegraphics[width=0.49\linewidth]{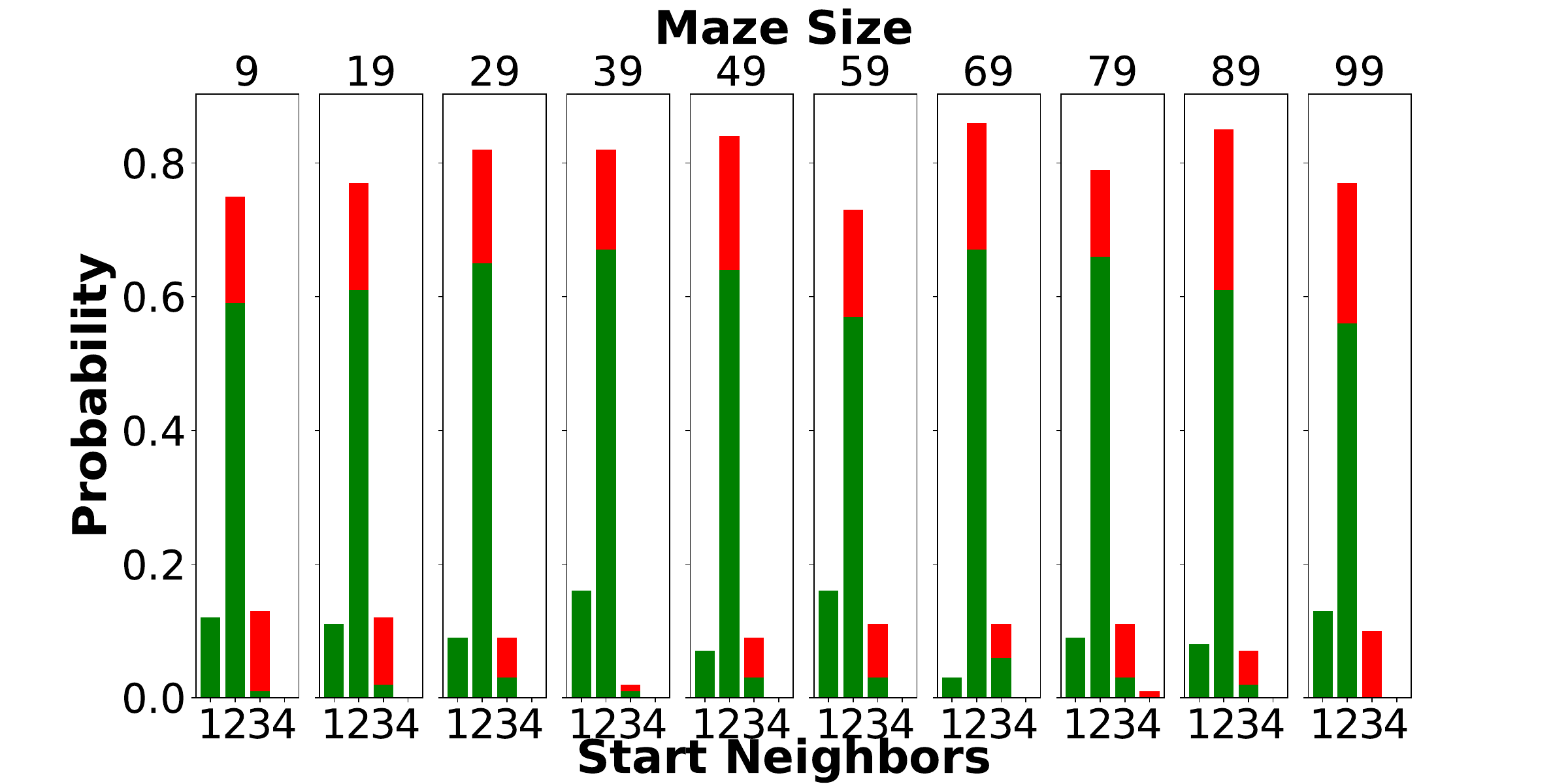}
        & \includegraphics[width=0.49\linewidth]{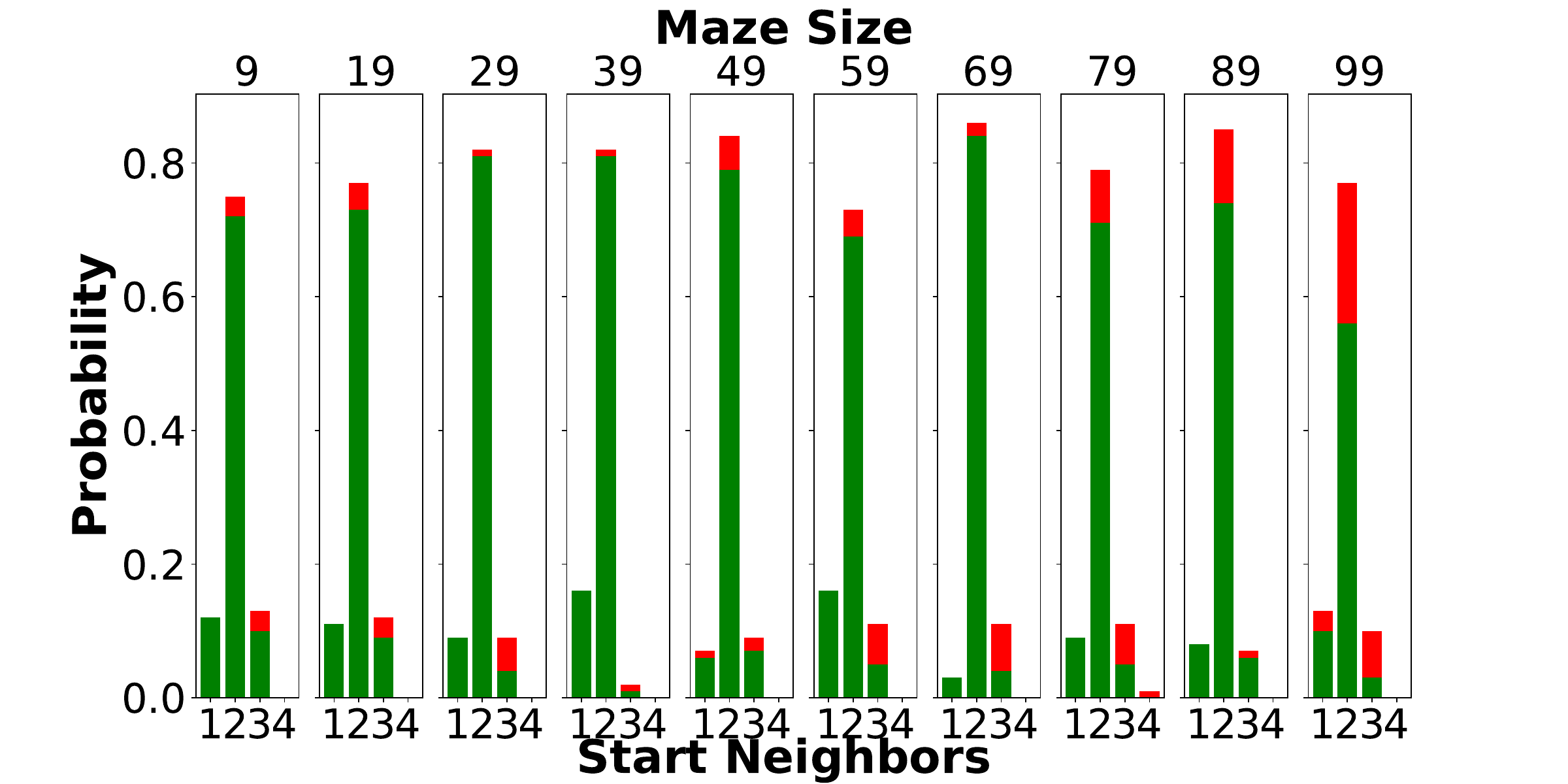}
    \end{tabular} \\
    \small
    \begin{tabular}{@{}ll@{}}
      \rule{10pt}{10pt} & Incorrect \\[0.5ex]
      \fbox{\rule{0pt}{10pt}\rule{10pt}{0pt}} & Correct
    \end{tabular}

    \caption{\small
      \DTNet and \PINet predictions for {\tt deadend\_start=False} mazes, split by maze size and number of start‐node neighbors. Accuracy drops as neighbor‐count rises.
    }
    \label{fig:extrap_no_deadend_split_neighbors}
\end{figure}

\reffig{fig:additional_test_acc} provides an alternative visualization of the test accuracy for the models of \refsec{sec:diversifying_training_data}, while \reffig{fig:extrap_no_deadend_split_neighbors} breaks down test accuracy by start node degree, when \texttt{deadend\_start} is set to {\tt True}. 

\begin{figure}
    \centering
    \includegraphics[width=0.95\linewidth]{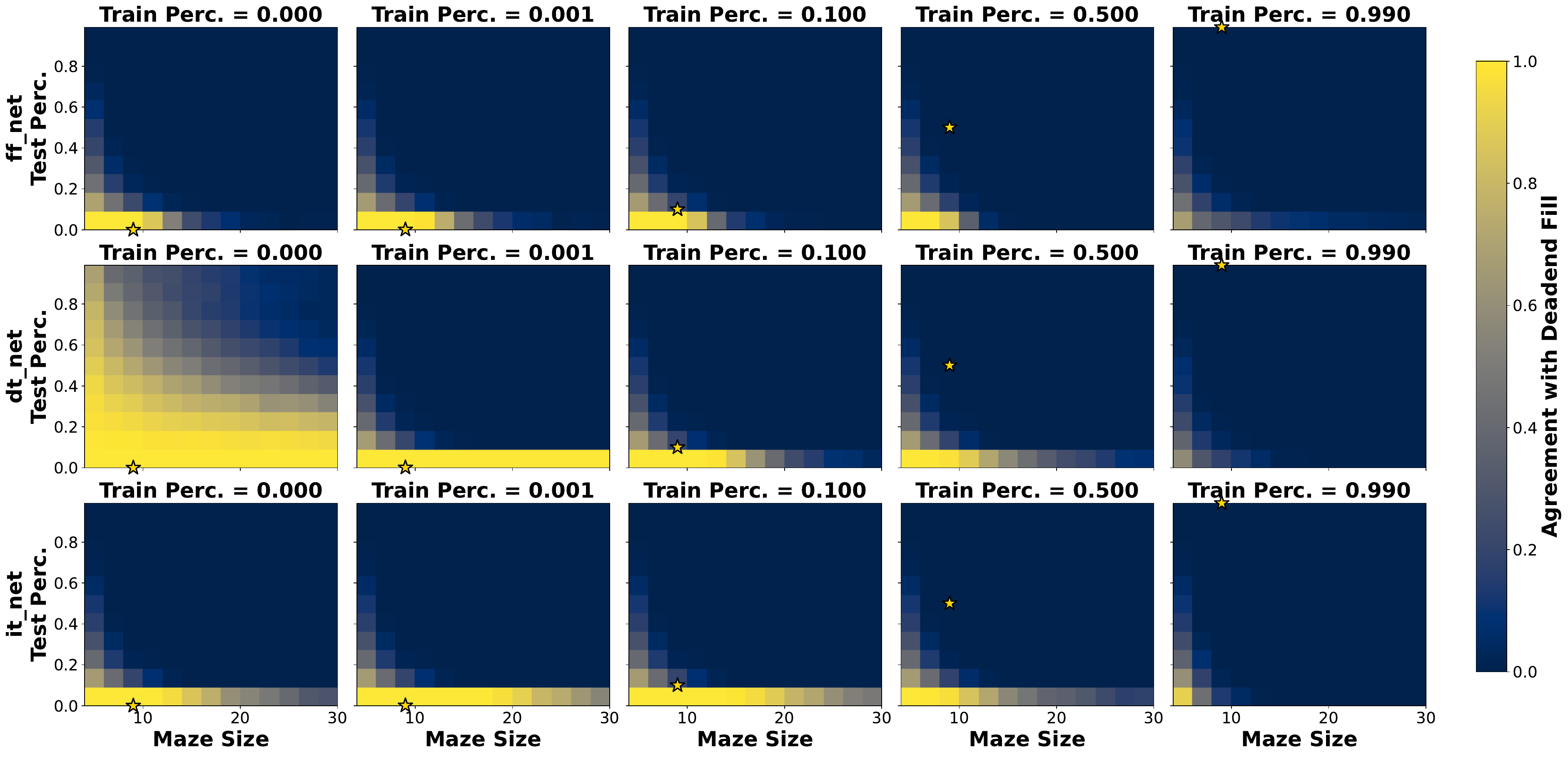}
    \caption{{\small Heatmap showing the agreement between the trained-from-scratch instances of \FFNet, \DTNet, and \ITNet with the deadend-fill algorithm, across various maze sizes and test percolation values.}}
    \label{fig:agreement_ff_dt_it_nets_deadend_fill}
\end{figure}

\begin{figure}
    \centering
    \includegraphics[width=0.95\linewidth]{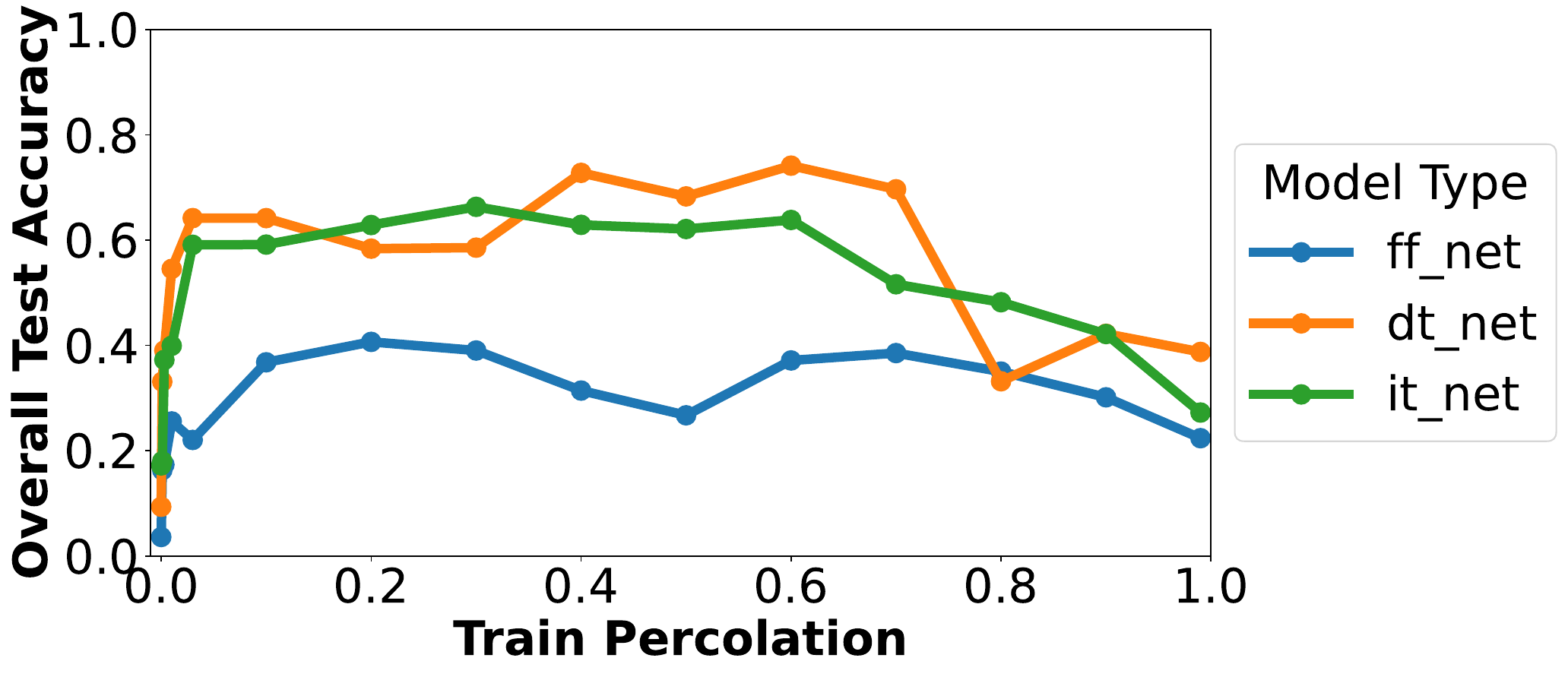}
    \caption{{\small Plots of overall test accuracy (average accuracy over all test mazes) as a function of train percolation for \DTNet, \ITNet, and \FFNet. Each point is a trained model.}}
    \label{fig:overall_test_accuracy}
\end{figure}

\section{The Ripser Wrapper}
\label{section:Pytorch-Ripser-wrapper}
A number of optimizations were applied to reduce the memory and compute time of TDA. Instead of providing a sequence of high-dimensional latent iterates directly to {\tt Ripser}, we use a smaller distance matrix containing pairwise distances between the iterates. This matrix is computed using an optimization from \citet{tralie2018quasi}, where the iterates are first compressed with singular value decomposition (SVD) to reduce memory costs. The SVD computation is performed in {\tt PyTorch} to leverage GPU acceleration. Initially, we also used a diagonal convolution optimization, also from \citet{tralie2018quasi}, to avoid redundant computations when constructing the distance matrix for the sliding window embedding. However, this second optimization was not used in our final TDA experiments, as we dropped the sliding window embedding.

\end{document}